\documentclass[11pt]{article}

\usepackage[preprint]{acl}

\usepackage{times}
\usepackage{latexsym}
\usepackage{amsmath}
\usepackage{amssymb}

\usepackage[T1]{fontenc}

\usepackage[utf8]{inputenc}

\usepackage{microtype}

\usepackage{inconsolata}

\usepackage{graphicx}

\usepackage{booktabs}
\usepackage{multirow}
\usepackage{graphicx}
\usepackage{amsmath}

\title{SeRV: Semantic-Aligned Residual Vector Quantization for American Sign Language Generation}

\author{
Hongyu Wu$^{1}$ \quad Xu Wu$^{1}$ \quad Tianhao Wu$^{2}$ \quad Jiawei Yu$^{2}$ \\
Phuc Nguyen$^{3}$ \quad Jian Liu$^{2}$ \quad Yi Wu$^{1}$ \\[0.4em]
$^{1}$University of Oklahoma \quad $^{2}$University of Georgia \\
$^{3}$University of Massachusetts Amherst
}

\usepackage{xspace}
\begin{document}

\newcommand{\methodName}{SeRV\xspace}
\newcommand{\tbd}[1]{#1}

\maketitle
\begin{abstract}
American Sign Language (ASL) generation remains challenging due to limited paired text-ASL motion data and the difficulty of learning motion representations both precise for reconstruction and predictable from linguistic input.
Existing methods rely on motion tokenizers optimized for reconstruction, without explicit semantic supervision from paired text.
As a result, the learned tokens remain limited in supporting semantically consistent and fine-grained ASL motion generation.
To address this limitation, we propose \textbf{SeRV} (\textbf{Se}mantic-Aligned \textbf{R}esidual \textbf{V}ector Quantization), a semantic-aligned RVQ tokenizer for ASL generation.
SeRV learns a semantically structured residual token space by combining global-level motion--text alignment with token-level text-conditioned supervision.
Building on this tokenizer, a Hierarchical GPT predicts residual motion tokens in a coarse-to-fine manner, generating structurally coherent and semantically aligned 3D ASL motion.
We further construct a large-scale reconstructed 3D ASL motion--text benchmark by recovering paired 3D motion from OpenASL videos.
Experiments on How2Sign and OpenASL show that SeRV outperforms six state-of-the-art baselines, with user studies assessing reconstruction quality and semantic fidelity. Samples of our reconstructed 3D ASL motion are available on our project page:\hspace{0.3em}\url{https://signvideos.github.io/Demo/}.

\end{abstract}

\section{Introduction}

\begin{figure}[t]
    \centering    \includegraphics[width=\columnwidth]{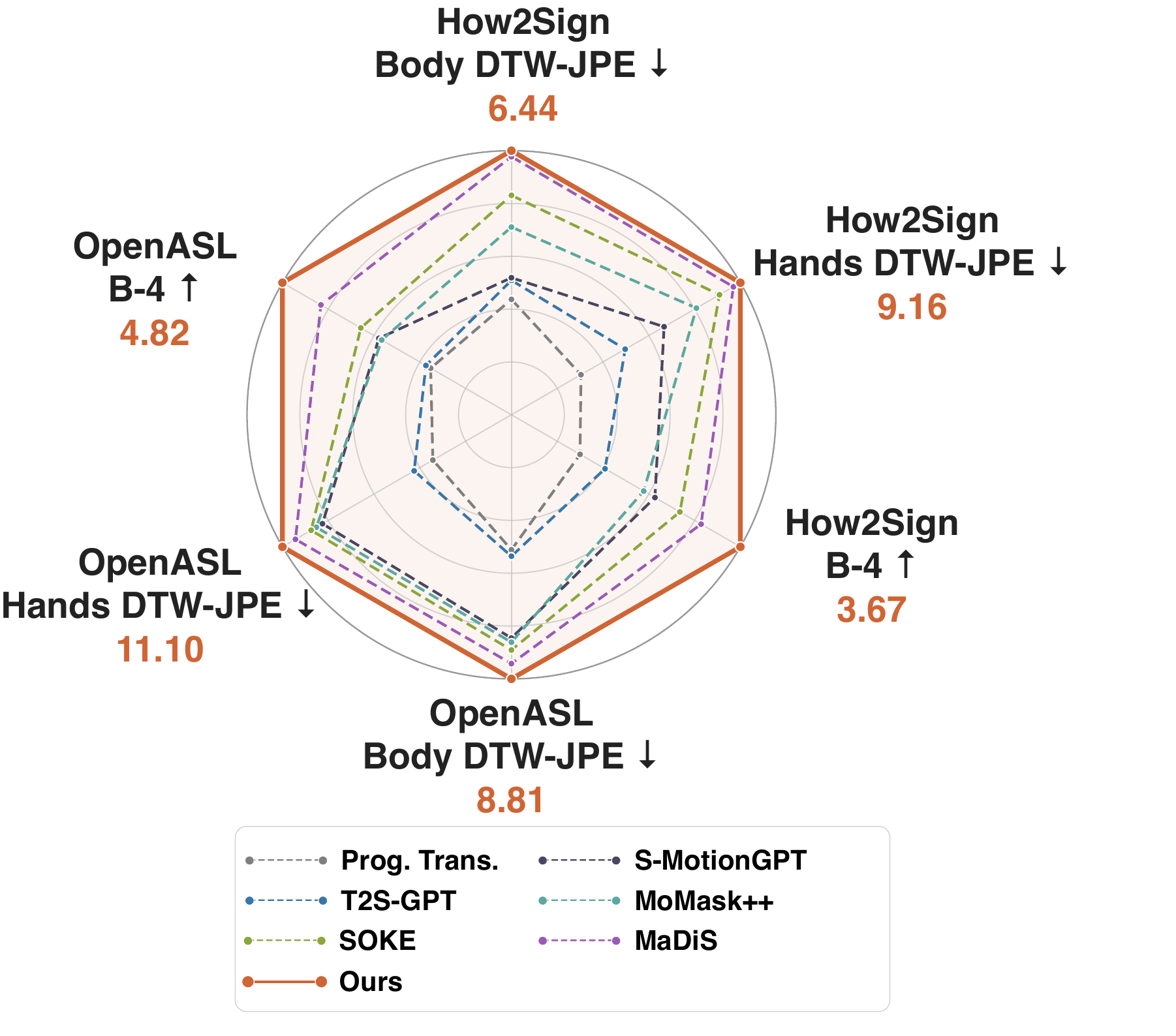}
    \caption{
    Comparison with state-of-the-art methods on How2Sign and OpenASL. Axes show body and hand DTW-JPE (lower is better) and back-translation BLEU-4 (B-4; higher is better), oriented so that farther from the center is better. \methodName achieves the best result on all six metrics.} 
    \label{fig:intro}
\end{figure}

\begingroup

American Sign Language (ASL) is the primary communication method of Deaf and Hard-of-Hearing (DHH) communities, but the lack of fluent signers creates substantial communication barriers. To bridge this gap, recent AI research has focused on Sign Language Translation (SLT)~\cite{jang2025lost,li2025uni,wong2024sign2gpt} and Sign Language Generation (SLG)~\cite{zuo2025signs,yu2024signavatars,fang2025signllm,yin2024t2s}.
Among SLG approaches, tokenization-based methods~\cite{yin2024t2s,zuo2025signs} convert continuous sign motions into discrete tokens, and some methods additionally use external sign dictionaries and retrieval to guide generation~\cite{zuo2025signs}.
Accurate generation also depends on whether the learned tokens preserve the fine-grained hand articulation that carries ASL meaning. Yet existing tokenizers are primarily trained for motion reconstruction, without explicit alignment between their token representations and the semantics of the conditioning text. Consequently, signs with distinct meanings may receive similar token representations, making the text-to-token mapping difficult to learn and potentially yielding natural-looking motions that convey the wrong meaning~\cite{zuo2025signs,yin2024t2s,zuo2026madis}. 

Moreover, current 3D SLG models are trained on small-scale data, as existing 3D ASL resources~\cite{duarte2021how2sign,yu2024signavatars} remain limited in scale and diversity, which restricts their generalization to open-domain vocabulary.

In this work, we propose \textbf{SeRV} (\textbf{Se}mantic-Aligned \textbf{R}esidual \textbf{V}ector Quantization), which trains a Residual Vector-Quantized Variational AutoEncoder (RVQ-VAE)~\cite{lee2022autoregressive,guo2024momask} with linguistic supervision rather than reconstruction alone, so that its residual codes become predictable from language.
SeRV trains the encoder with two semantic alignment objectives: a global-level alignment on the base-level (coarsest) tokens, and a token-level motion--text alignment on the fully quantized latent, driving coarse tokens to encode sentence-level semantics while finer residuals refine articulatory details.
Building on this token space, we further propose Hierarchical GPT, an RVQ-aware autoregressive generator that predicts tokens level by level, conditioned on per-token text features and the previously decoded lower-level motion tokens. Unlike flat~\cite{guo2025snapmogen} or two-stage~\cite{guo2024momask} prediction, this design preserves the residual hierarchy and yields structurally coherent ASL motion.
Our contributions are summarized as follows:
\begin{itemize}
\item We propose \textbf{SeRV}, a semantic-aligned RVQ tokenizer whose residual tokens are precise for reconstruction and predictable from text, and the Hierarchical GPT generator.
\item We construct a reconstructed 3D ASL motion--text benchmark by recovering paired 3D sign motion from OpenASL videos using the official paired clips and data split.
\item Experiments on How2Sign and OpenASL show that SeRV outperforms six state-of-the-art baselines on all six reported metrics (Figure~\ref{fig:intro}; Table~\ref{tab:main_results}). User studies assess reconstruction quality and the semantic fidelity of generated motion.
\end{itemize}
\endgroup

\section{Related Work}
\textbf{Sign Language Generation (SLG).}

Early  SLG methods often adopt cascaded pipelines, bridging text and sign motion via glosses, skeletons, or pose sequences~\cite{stoll2020text2sign,zelinka2020neural}. These intermediate representations provide linguistic or structural guidance but require costly annotations and risk information bottlenecks or error accumulation~\cite{saunders2021continuous,zuo2025signs,zuo2026madis}. Others generate continuous sign poses or videos directly from text via Transformers~\cite{saunders2020progressive,saunders2021continuous,saunders2022signing,fang2025signllm}, GANs~\cite{stoll2020text2sign}, or diffusion models~\cite{fang2025signdiff,qi2024signgen}. However, operating in a continuous output space makes long-horizon generation harder to stabilize and offers limited discrete structure for scalable autoregressive modeling.
Recent SLG methods tokenize 3D sign motion into discrete codes. Early efforts use pose-level VQ-VAE~\cite{xie2024g2p} or VQ-GAN-style video tokenizers~\cite{esser2021taming,xie2024sign}. More recent work targets 3D motion: SignAvatars~\cite{yu2024signavatars} pairs a large-scale 3D holistic sign dataset with a VQ-VAE codebook; T2S-GPT~\cite{yin2024t2s} uses dynamic vector quantization for adaptive-length codes; SOKE~\cite{zuo2025signs} tokenizes upper-body, left-hand, and right-hand separately and generates multilingual signs via a pretrained LM with dictionary retrieval; MaDiS~\cite{zuo2026madis} combines part-wise tokens with masked diffusion language modeling and tri-level cross-modal pretraining.

Despite these advances, existing motion tokenizers primarily preserve motion fidelity through reconstruction-oriented objectives, without encouraging the learned representations to capture the linguistic semantics of signs.
Consequently, tokens that accurately reconstruct gestures may not be well aligned with their textual meanings.

\textbf{3D ASL Resources.}
Existing resources such as How2Sign~\cite{duarte2021how2sign}, SignAvatars~\cite{yu2024signavatars}, and Neural Sign Actors~\cite{baltatzis2024neural} provide valuable 3D sign resources, but paired 3D ASL motion--text data remain limited in scale and diversity.
OpenASL~\cite{shi2022open} offers a large collection of in-the-wild continuous ASL videos with paired text, but lacks 3D motion annotations.
We reconstruct SMPL-X motion from OpenASL videos to expand paired 3D ASL motion--text resources and support larger-scale training of sign generation models.

\section{Methodology}

\begin{figure*}[th]
    \centering
    \includegraphics[width=1.0\linewidth]{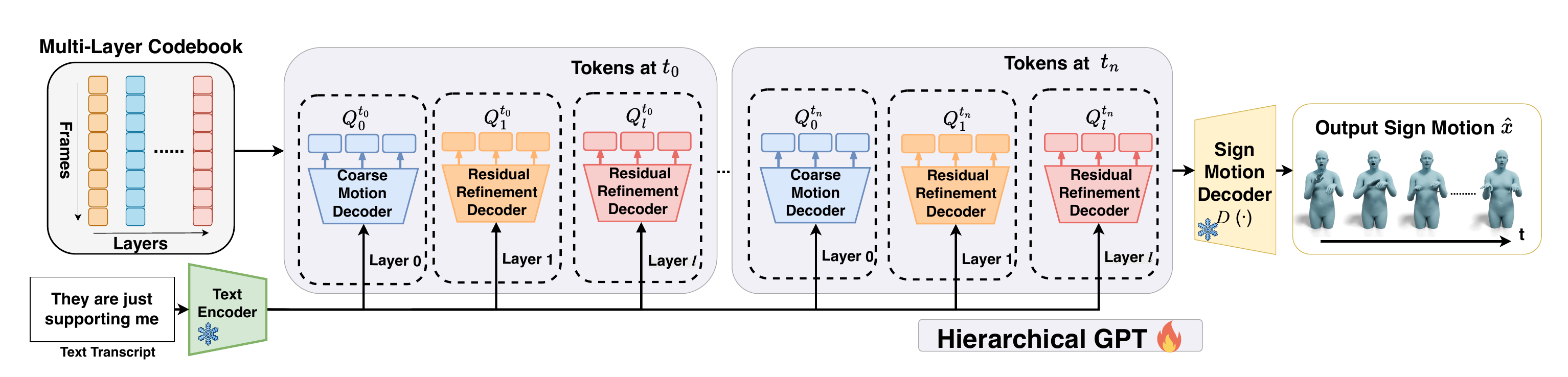}
    \caption{
    Hierarchical GPT for coarse-to-fine ASL motion generation. The diagram labels $Q_l^t$ denote discrete tokens $k_t^{(l)}$; output $\hat{x}$ denotes $\hat{\mathbf{S}}$.
    Given the text feature from the frozen condition encoder, the generator
    autoregressively predicts SeRV tokens over time and RVQ levels. For each time
    step, a coarse decoder first predicts the level-$0$ token, while residual
    decoders progressively predict higher-level tokens conditioned on the temporal
    history, the generated coarser tokens, and the text feature. The predicted
    multi-level tokens are then decoded by the shared motion decoder to reconstruct
    the ASL motion sequence. 
    }
    \label{fig:hgpt}

\end{figure*}

The overview of our framework is shown in Figure~\ref{fig:rvq} and Figure~\ref{fig:hgpt}. Our method has two stages: \textbf{SeRV} (\textbf{Se}mantic-Aligned \textbf{R}esidual \textbf{V}ector Quantization), a text-aligned RVQ tokenizer, and Hierarchical GPT, a coarse-to-fine token generator. SeRV converts continuous 3D ASL motion into residual discrete tokens while aligning the motion representations with the paired transcript. Hierarchical GPT then predicts these residual tokens from text in a coarse-to-fine autoregressive order, and the predicted tokens are decoded by the learned motion decoder to reconstruct the final 3D ASL motion.

\subsection{Preliminaries of RVQ-VAE}
\label{sec:preliminaries}
Given a sign motion sequence $\mathbf{S}\in\mathbb{R}^{T\times d_s}$, with $T$ frames and motion-feature dimension $d_s$, a motion encoder $\mathcal{E}$ produces a continuous latent sequence $\mathbf{z}=\mathcal{E}(\mathbf{S})\in\mathbb{R}^{T'\times C}$, with temporal length $T'$ and feature dimension $C$.
A codebook $\mathcal{C}=\{\mathbf{e}_m\}_{m=0}^{K-1}$ is a learnable set of $K$ embedding vectors $\mathbf{e}_m\in\mathbb{R}^{C}$; each latent vector is quantized to its nearest code, and the selected index serves as a discrete token.
We adopt residual vector quantization with $L$ codebooks $\{\mathcal{C}_l\}_{l=0}^{L-1}$~\cite{guo2024momask,lee2022autoregressive}.
Starting from $\mathbf{r}^{(0)}=\mathbf{z}$, stage $l$ computes $\mathbf{q}^{(l)}=Q^{(l)}(\mathbf{r}^{(l)})$ and updates the residual $\mathbf{r}^{(l+1)}=\mathbf{r}^{(l)}-\mathbf{q}^{(l)}$, where $Q^{(l)}(\cdot)$ is nearest-neighbor quantization with $\mathcal{C}_l$.
The selected index $k_t^{(l)}\in\{0,\ldots,K-1\}$ at position $t$ is the discrete motion token. Each residual $\mathbf{r}^{(l)}$ and quantized sequence $\mathbf{q}^{(l)}$ lies in $\mathbb{R}^{T'\times C}$.
The final quantized latent is $\hat{\mathbf{z}}=\sum_{l=0}^{L-1}\mathbf{q}^{(l)}$, decoded to $\hat{\mathbf{S}}=\mathcal{D}(\hat{\mathbf{z}})$ by the motion decoder $\mathcal{D}$.
The RVQ-VAE is trained with a reconstruction loss and a per-level embedding constraint:
\begin{equation}
\mathcal{L}_{\mathrm{RVQ}}
=
\mathcal{L}_{\mathrm{rec}}
+
\beta\sum_{l=0}^{L-1}\left\|\mathbf{r}^{(l)}-\mathrm{sg}\!\left[\mathbf{q}^{(l)}\right]\right\|_F^2,
\end{equation}
where $\mathcal{L}_{\mathrm{rec}}=\|\mathbf{S}-\hat{\mathbf{S}}\|_1$, $\mathrm{sg}[\cdot]$ is the stop-gradient operator, and $\beta$ weights the embedding constraint. Here, $\|\cdot\|_F$ is the Frobenius norm over the latent sequence, and the reconstruction $\ell_1$ norm sums absolute elementwise errors.
Codebooks receive no gradients; each code is instead updated as an exponential moving average (decay $0.99$) of the residuals assigned to it, and unused codes are reset to residuals sampled from the current batch.
We apply a straight-through estimator at every level, so the reconstruction and semantic losses update the encoder through the quantizer.

\subsection{\methodName Model Architecture}
\label{sec:stage1}

\begin{figure*}[!t]
    \centering 
    \begingroup
    \setlength{\unitlength}{\textwidth}
    \begin{picture}(1,0.39552)
    \put(0,0){\includegraphics[width=\textwidth]{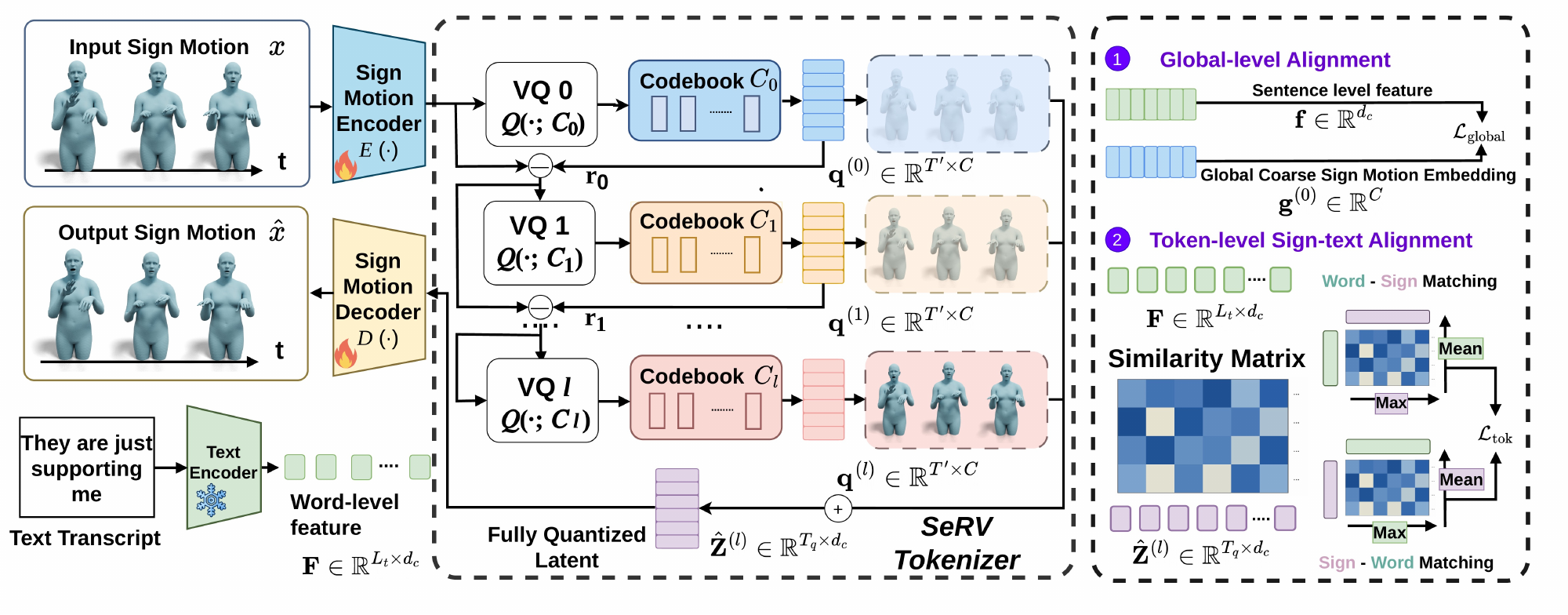}}

    \end{picture}
    \endgroup
    
    \caption{
    Overview of \methodName. In the diagram, $x$ and $\hat{x}$ denote $\mathbf{S}$ and $\hat{\mathbf{S}}$, respectively. The late-interaction panel shows the projected features $\tilde{\mathbf{F}}$ and $\tilde{\mathbf{Z}}$.     We introduce a global-level loss $\mathcal{L}_{\mathrm{global}}$ on the level-0 motion representation and a token-level loss $\mathcal{L}_{\mathrm{tok}}$ on the fully quantized latent sequence. Together with the standard RVQ tokenizer objective $\mathcal{L}_{\mathrm{RVQ}}$, these objectives encourage the learned discrete motion tokens to preserve motion fidelity while retaining text-relevant semantics.}

    \label{fig:rvq} 
\end{figure*}

Unlike a reconstruction-only RVQ-VAE, \methodName adds semantic
supervision from paired transcripts during tokenizer training.
As illustrated in Figure~\ref{fig:rvq}, a frozen T5 encoder~\cite{chung2024scaling} provides both sentence-level and token-level text features from the same transcript. We align the sentence-level embedding with the level-0 quantized latent $\mathbf{q}^{(0)}$, a coarse content
representation of the ASL motion, and use the token-level features in a
motion--text late interaction loss over quantized motion latents. These losses
inject global and token-level textual supervision into the tokenizer.
\paragraph{Global-Level Sign--Text Alignment.}

We align $\mathbf{q}^{(0)}$ with the paired
text representation so that the learned ASL motion tokens retain
text-relevant semantics. Given the paired transcript $\mathbf{c}$, a frozen pretrained T5 encoder~\cite{chung2024scaling} $\mathcal{E}_{\mathrm{text}}$ produces token-level text features $\mathbf{F}=\mathcal{E}_{\mathrm{text}}(\mathbf{c})\in\mathbb{R}^{L_t\times d_c}$, where $L_t$ is the tokenized text length and $d_c$ is the T5 hidden dimension.
We use the average hidden state over non-padded tokens as the
sentence-level representation $\mathbf{f}\in\mathbb{R}^{d_c}$. On the motion
side, we summarize $\mathbf{q}^{(0)}$ into a global motion embedding
$\mathbf{g}^{(0)}$ using temporal mean pooling. The motion and text embeddings
are then projected into a shared semantic space and $\ell_2$-normalized.

Let $\hat{\mathbf{g}}_i^{(0)}$ and $\hat{\mathbf{f}}_j$ denote these embeddings for samples $i$ and $j$ in a mini-batch of size $B$, with cosine similarity $s_{ij}^{(0)}=\mathrm{sim}(\hat{\mathbf{g}}_i^{(0)},\hat{\mathbf{f}}_j)$. $\mathcal{L}_{\mathrm{global}}$ is a symmetric InfoNCE loss~\cite{oord2018representation,radford2021learning} over $s^{(0)}$ with temperature $\tau$, averaging the motion-to-text and text-to-motion directions; matched motion--text pairs serve as positives and the other in-batch samples as negatives.

\paragraph{Token-Level Sign--Text Alignment.}

The global-level contrastive loss provides global alignment but leaves token-level text--motion interactions implicit. We therefore introduce a late-interaction loss~\cite{yao2021filip} that averages token-wise maximum similarities between text tokens and quantized ASL motion latents.

For text features, we reuse the T5 token hidden states $\mathbf{F}\in\mathbb{R}^{L_t\times d_c}$; for motion features, we use the fully quantized latent $\hat{\mathbf{z}}=\sum_{l=0}^{L-1}\mathbf{q}^{(l)}\in\mathbb{R}^{T'\times C}$. Separate token-level projection heads map them into a shared normalized space, with projection dimension $d_a$, producing $\tilde{\mathbf{F}}\in\mathbb{R}^{L_t\times d_a}$ and $\tilde{\mathbf{Z}}\in\mathbb{R}^{T'\times d_a}$. For each ASL motion--text pair, we compute a token-wise cosine
similarity matrix $\mathbf{A}\in\mathbb{R}^{L_t\times T'}$,
where
$\mathbf{A}_{u,v}
=
\mathrm{sim}(\tilde{\mathbf{F}}_{u},\tilde{\mathbf{Z}}_{v})$
measures the similarity between the $u$-th text token and the
$v$-th motion latent. 
Let $\Omega_F$ and $\Omega_Z$ denote valid text-token and motion-latent
positions. We define the late-interaction scores as
$
s_{F\to Z}
=
\frac{1}{|\Omega_F|}
\sum_{u\in\Omega_F}
\max_{v\in\Omega_Z}
\mathbf{A}_{u,v}$ 
and
$s_{Z\to F}
=
\frac{1}{|\Omega_Z|}
\sum_{v\in\Omega_Z}
\max_{u\in\Omega_F}
\mathbf{A}_{u,v}.
$
Each valid token is therefore matched to its closest token in the other
modality.

Given a mini-batch of size $B$, let $s_{ij}^{Z\to F}$ denote the motion-to-text score obtained by averaging over the motion latents of sample $i$ and maximizing over the text tokens of sample $j$.
Conversely, $s_{ij}^{F\to Z}$ averages over the text tokens of sample $i$ and maximizes over the motion latents of sample $j$.
Thus, the row index is the query sample and the column index is the candidate sample in each direction.

$\mathcal{L}_{\mathrm{tok}}$ applies the same symmetric InfoNCE loss to these scores with a token-level temperature $\tau'$.

\paragraph{Training Objective.}

The overall training objective of \methodName{} is defined as
\begin{equation}
\mathcal{L}_{\mathrm{SeRV}}
=
\mathcal{L}_{\mathrm{RVQ}}
+
\lambda_{\mathrm{global}}\mathcal{L}_{\mathrm{global}}
+
\lambda_{\mathrm{tok}}\mathcal{L}_{\mathrm{tok}} .
\end{equation}
Here, $\mathcal{L}_{\mathrm{RVQ}}$ denotes the standard RVQ tokenizer objective,
$\mathcal{L}_{\mathrm{global}}$ aligns the coarse level-0 motion representation
with sentence-level text semantics, while $\mathcal{L}_{\mathrm{tok}}$
encourages fine-grained interactions between text tokens and quantized ASL
motion latents. The coefficients $\lambda_{\mathrm{global}}$ and
$\lambda_{\mathrm{tok}}$ balance the two semantic alignment objectives.

\subsection{Hierarchical GPT}
\label{sec:stage2}

\paragraph{Hierarchy-aligned generation.}

We propose Hierarchical GPT, an RVQ-aware autoregressive
generator that predicts SeRV tokens over both time and quantization depth.
We reuse the frozen T5 encoder and its token-level text features
$\mathbf{F}=\mathcal{E}_{\mathrm{text}}(\mathbf{c})\in\mathbb{R}^{L_t\times d_c}$, where
$\mathbf{c}$ is the input text. The sentence-level vector $\mathbf{f}$ used for global alignment is the pooled representation of $\mathbf{F}$. Let
$\mathbf{k}_{1:T_q}^{0:L-1}$ be the multi-level token hierarchy produced by
\methodName, where $L$ is the number of RVQ levels, $T_q$ is the token
sequence length ($T_q=T'$ for the motion tokens), and $k_t^{(l)}$ is the token at time $t$ and level $l$.
Following the text-aligned hierarchy learned by \methodName, the generator
predicts lower-level coarse tokens before higher-level residual tokens:
\begin{equation}
p(\mathbf{k}_{1:T_q}^{0:L-1}\mid \mathbf{F})
=
\prod_{t=1}^{T_q}\prod_{l=0}^{L-1}
p(
k_t^{(l)}
\mid
\mathbf{k}_{<t}^{0:L-1},
\mathbf{k}_{t}^{0:l-1},
\mathbf{F}
)
\label{eq:hgpt_factor}
\end{equation}
where $p(\cdot)$ is the categorical distribution over codebook entries,
$\mathbf{k}_{<t}^{0:L-1}$ denotes all tokens before time $t$, and
$\mathbf{k}_{t}^{0:l-1}$ denotes coarser tokens already generated (empty for $l=0$). This keeps temporal causality and coarse-to-fine
dependencies.

\paragraph{Coarse-to-fine decoding.}
We implement Eq.~\eqref{eq:hgpt_factor} with a coarse token decoder followed by residual token decoders. At each time step $t$, the coarse decoder predicts the level-0 token $k_t^{(0)}$ from the previous token history and the text feature $\mathbf{F}$, capturing the coarse text-conditioned motion pattern. The coarse decoder is a causal Transformer built with LLaMA-style blocks, including RMSNorm, SwiGLU, and rotary positional embeddings (RoPE)~\cite{touvron2023llama}.

For each higher level $l>0$, a residual decoder predicts $k_t^{(l)}$ conditioned on the previous token history $\mathbf{k}_{<t}^{0:L-1}$, the coarser tokens $\mathbf{k}_{t}^{0:l-1}$ already generated at the current time step, and the text feature $\mathbf{F}$. 

The embeddings of these coarser tokens are fed to the residual decoder, allowing higher levels to refine the current coarse representation rather than generate the motion state from scratch. Since residual decoders mainly model refinement details, we use fewer Transformer layers while retaining attention layers for fine-grained articulation modeling. This decoding process follows the coarse-to-fine hierarchy learned by \methodName.

\paragraph{Training objective.}
Let $\hat{\mathbf{y}}_t^{(l)}$ denote the predicted token distribution at
level $l$ and time step $t$, and let $k_t^{(l)}$ be the corresponding
ground-truth token from the frozen \methodName tokenizer. The generator is
trained with a level-wise cross-entropy loss:
\begin{equation}
\mathcal{L}_{\mathrm{HGPT}}
=
\sum_{l=0}^{L-1}
\lambda_l
\frac{1}{T_q}
\sum_{t=1}^{T_q}
\operatorname{CE}
\bigl(
\hat{\mathbf{y}}_t^{(l)},
k_t^{(l)}
\bigr),
\end{equation}
where $T_q$ is the token length, $\lambda_l$ is the loss weight for
level $l$, and $\operatorname{CE}(\cdot,\cdot)$ denotes cross-entropy.

\section{Performance Evaluation}

\begin{table*}[t]
\centering
\caption{
Comparison with state-of-the-art methods on How2Sign and OpenASL.
Body and Hands are DTW-JPE (lower is better); B-4 denotes back-translation BLEU-4 (higher is better).
}
\label{tab:main_results}
\scriptsize
\setlength{\tabcolsep}{4pt}
\renewcommand{\arraystretch}{0.85}
\resizebox{0.8\textwidth}{!}{
\begin{tabular}{lcccccc}
\toprule
\multirow{2}{*}{Method}& \multicolumn{3}{c}{How2Sign}& \multicolumn{3}{c}{OpenASL} \\\cmidrule(lr){2-4}\cmidrule(lr){5-7}
& Body$\downarrow$ & Hands$\downarrow$ & B-4$\uparrow$
& Body$\downarrow$ & Hands$\downarrow$  & B-4$\uparrow$ \\
\midrule

Prog. Trans.$^{\ddagger}$~\cite{saunders2020progressive}& 14.74 & 30.17 & 1.10
& 17.23 & 32.35 & 1.70 \\

T2S-GPT$^{\ddagger}$~\cite{yin2024t2s}& 12.65 & 18.44 & 1.50
& 16.45 & 26.13 & 1.80 \\

S-MotionGPT$^{*}$~\cite{jiang2023motiongpt}& 12.41 & 13.74 & 2.30
& 10.42 & 13.45 & 2.79 \\

MoMask++$^{*}$~\cite{guo2025snapmogen}& 9.06 & 11.34 & 2.12
& 10.23 & 13.03 & 2.73 \\

SOKE$^{\dagger}$~\cite{zuo2025signs}& 7.75 & 10.08 & 2.70
& 9.89 & 12.69 & 3.17 \\

MaDiS~\cite{zuo2026madis}& 6.59 & 9.45 & 3.04
& 9.35 & 11.77 & 4.01 \\
\midrule

\textbf{SeRV}& \textbf{6.44} & \textbf{9.16} & \textbf{3.67}
& \textbf{8.81} & \textbf{11.10} & \textbf{4.82} \\

\bottomrule
\end{tabular}
}

\begin{minipage}{0.98\textwidth}
\scriptsize
\emph{Notes.} Hands averages both hands.
$^{*}$: adapted motion-generation models;
$^{\ddagger}$: How2Sign DTW-JPE from SOKE reimplementations;
$^{\dagger}$: external sign dictionaries.
All B-4 scores are re-evaluated using a back translator; all OpenASL results are reproduced in this work. See Sections~\ref{subsec:experimental_methodology}--\ref{subsec:sota_comparison} and Appendix~\ref{subsec:metrics} for sources and evaluation details.
\end{minipage}

\end{table*}

\subsection{Experimental Methodology}
\label{subsec:experimental_methodology}
\paragraph{Datasets.}
We evaluate on two ASL datasets, How2Sign~\cite{duarte2021how2sign} and OpenASL~\cite{shi2022open}. For How2Sign, we use its SMPL-X motion annotations and the official train/validation/test split. 
For OpenASL, we reconstruct 3D SMPL-X signing motion from the source videos; the reconstruction pipeline and dataset statistics are detailed in Appendix~\ref{sec:pose_curation} and~\ref{app:dataset_comparison}. For both datasets, we represent sign motion as sequences of SMPL-X parameters covering upper-body and hand joints. We assess reconstruction quality in Appendix~\ref{app:reconstruction_audit}.

\paragraph{Evaluation metrics \& Implementation.}
Since generated sign sequences may differ in length from the ground truth, we follow prior 3D sign language generation work and use Dynamic Time Warping over Joint Position Error (DTW-JPE)~\cite{zuo2025signs,cai2019dtwnet} to measure sequence-level motion distance.
DTW temporally aligns the generated and reference sequences before computing joint-position errors.
We compute DTW-JPE separately for the body and each hand, and report Body and Hands, where Hands is the mean of the left- and right-hand errors.
For tokenizer reconstruction, where the generated motion is frame-aligned with the input, temporal alignment is unnecessary; we instead report PA-MPJPE, the Mean Per-Joint Position Error after Procrustes alignment with rotation, translation, and scale. 
We use a pre-trained back-translator~\cite{li2025uni} and report BLEU-4 (B-4); details of the back-translator and formal metric definitions are provided in Appendix~\ref{subsec:metrics}.
We train the framework in two stages. \methodName uses $L=3$ RVQ levels,
latent dimension $C=512$, and codebook size $K=512$ per level. Hierarchical GPT uses a
hidden dimension of $1024$ and $16$ attention heads; the coarse decoder and the two residual decoders contain $9$, $6$, and $3$ Transformer layers, respectively. Full details are in Appendix~\ref{sec:implementaion_details}.

\begin{figure*}[!t]
    \centering 
    \includegraphics[width=0.95\textwidth,height=1\textheight,keepaspectratio]{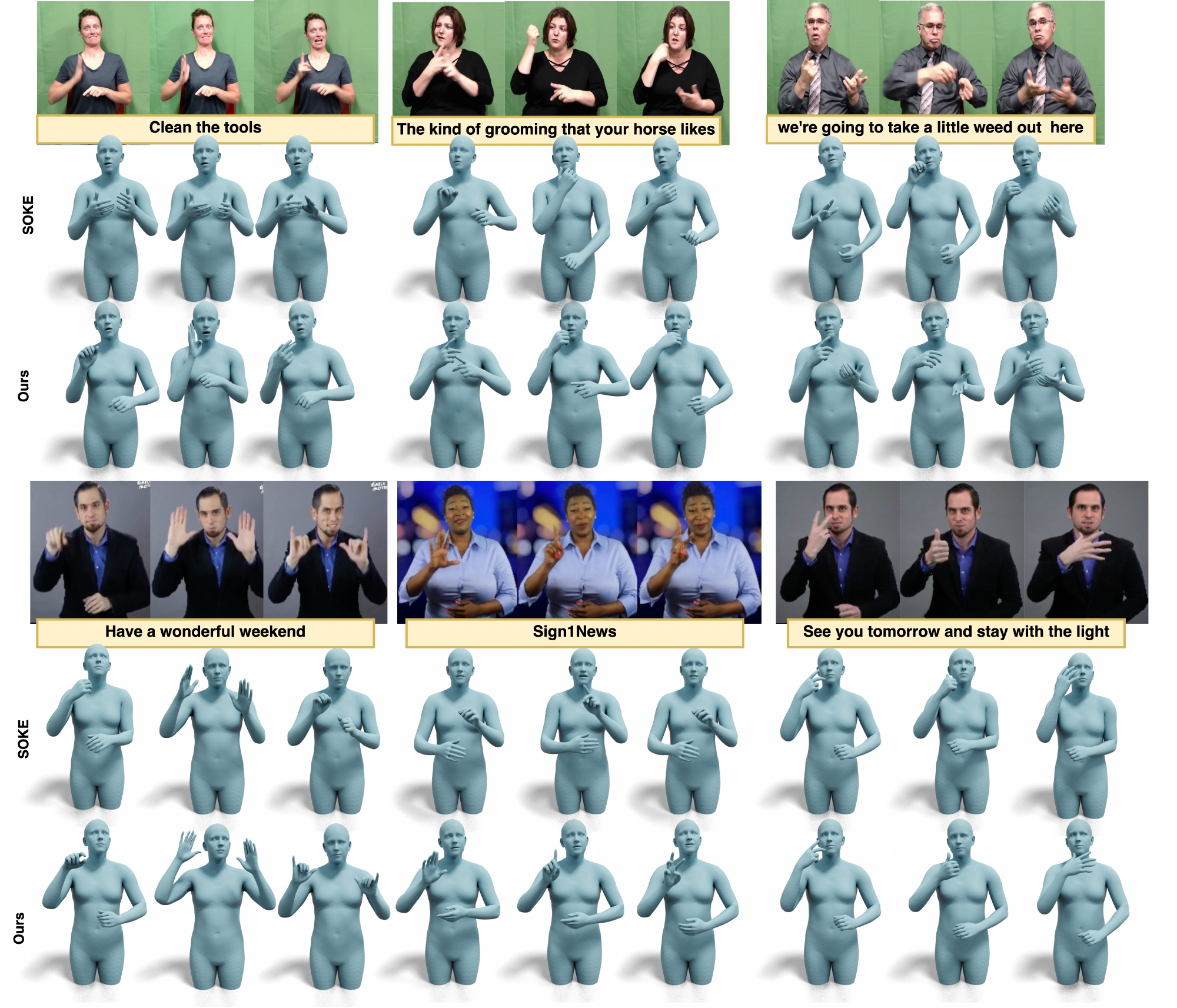}
    
    \caption{Qualitative comparison of text-to-3D ASL generation. For each of six examples, we show the reference signer video (top) and the 3D motion generated by SOKE and by our method (\methodName). The top three examples are from How2Sign; the bottom three are from OpenASL. More qualitative example videos are available on our project page: \url{https://signvideos.github.io/Demo/}.
 }
 
    \label{fig:qualitative_results}
\end{figure*}

\subsection{Comparison with State-of-the-Art Methods}
\label{subsec:sota_comparison}

We compare \methodName with three baseline groups (Table~\ref{tab:main_results}): sign-generation models Progressive Transformers~\cite{saunders2020progressive} and T2S-GPT~\cite{yin2024t2s}; general motion-generation models S-MotionGPT~\cite{jiang2023motiongpt} and MoMask++~\cite{guo2025snapmogen}, adapted to SMPL-X sign motion; and recent sign-specific models SOKE~\cite{zuo2025signs} and concurrent MaDiS~\cite{zuo2026madis}.
How2Sign DTW-JPE values are taken from prior work~\cite{zuo2025signs,zuo2026madis}, including SOKE's reimplementations of Prog.\ Trans.\ and T2S-GPT.
All OpenASL results come from our implementations and reproductions under the same SMPL-X representation, data split, and evaluation protocol.
As shown in Table~\ref{tab:main_results}, earlier sign-generation methods and adapted general motion generators trail recent sign-specific models, especially on hand motion, suggesting that tokenizers designed for general motion struggle to capture fine-grained sign articulation.
\methodName achieves the best results on all metrics on both How2Sign and OpenASL.
Notably, it outperforms SOKE without relying on an external sign dictionary.

\paragraph{Performance of Tokenizer Reconstruction.}
We further evaluate whether the proposed tokenizer preserves the input 3D motion.

As shown in Table~\ref{tab:reconstruction_results}, \methodName achieves the lowest reconstruction errors on both How2Sign and OpenASL.
Compared with the standard VQ-VAE and the part-wise Multi-VAE tokenizer used in SOKE, \methodName consistently reduces body and hand reconstruction errors, indicating that the proposed semantic alignment does not compromise motion fidelity.
The gains are especially important for downstream generation, as a more accurate residual token space provides cleaner targets for predicting both coarse body motion and fine-grained hand articulation.

\begin{table}[t]
\centering
\caption{Reconstruction performance of different motion tokenizers on How2Sign and OpenASL. We report PA-MPJPE in mm (lower is better).} 
\label{tab:reconstruction_results}
\resizebox{\columnwidth}{!}{
\begin{tabular}{lcccc}
\toprule
\multirow{2}{*}{Method}
& \multicolumn{2}{c}{How2Sign}
& \multicolumn{2}{c}{OpenASL} \\
\cmidrule(lr){2-3}
\cmidrule(lr){4-5}
& Body$\downarrow$
& Hands$\downarrow$
& Body$\downarrow$
& Hands$\downarrow$ \\
\midrule
VQ-VAE
& 24.07 & 7.96
& 28.90 & 7.45 \\

Multi-VAE
& 19.37 & 6.65
& 24.60 & 6.22 \\

\textbf{\methodName}
& \textbf{16.23} & \textbf{6.13 }
& \textbf{17.53} & \textbf{5.65} \\
\bottomrule
\end{tabular}
 }

\end{table}

\subsection{Qualitative Comparison}

We provide qualitative comparisons in Figure~\ref{fig:qualitative_results}.
For each input transcript, we show the 3D signing motion generated by SOKE and \methodName.
Compared with SOKE, \methodName produces more accurate hand configurations, especially for the fine-grained handshapes that carry important linguistic information; for instance, in the \textit{``Clean the tools''} example, the baseline outputs leave the hands in a weakly articulated, near-static pose, whereas our method reproduces the intended handshape.
Our method also yields more plausible upper-body posture with fewer self-contact artifacts.

\subsection{User Study}
\label{sec:user_study}

In addition to the objective metrics presented in the previous
tables, we conducted a user study involving 5 professional ASL signers to evaluate the
semantic alignment between generated sign motions and their
corresponding text annotations.

\begin{figure}[t]
    \centering
    \includegraphics[width=0.9\linewidth]{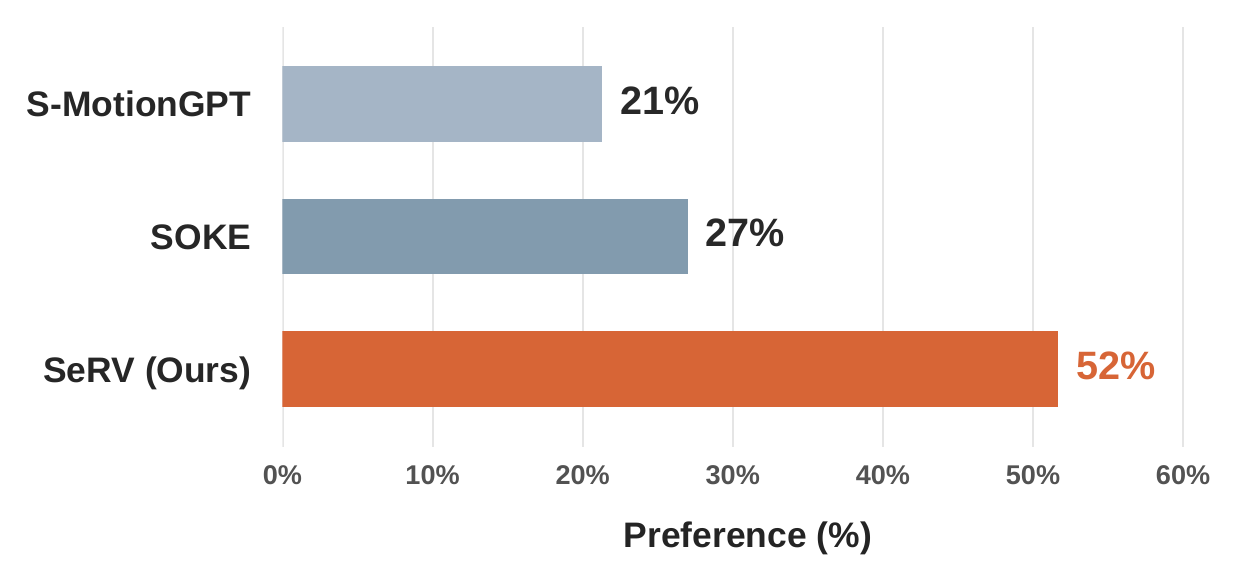}
    \caption{
        Preference rates from 5 professional ASL signers selecting the method that best conveys text meaning.
    }
    \label{fig:user_study}
\end{figure}

Specifically, we randomly selected 20 sign sequences and presented
participants with avatars generated by different methods
alongside their corresponding text annotations.
For each sequence, participants were asked to select the
generated avatar that best conveyed the semantic meaning
of the given text.
The study was approved by our institution's IRB (Appendix~\ref{app:user_study_details}).
MaDiS~\cite{zuo2026madis} is not included because it was open-sourced after the study had been distributed to participants.
As shown in Figure~\ref{fig:user_study}, our method was selected in 52\% of the trials, substantially surpassing SOKE at 27\% and S-MotionGPT at 21\%.
These results align with the above qualitative and quantitative evaluations, indicating that \methodName generates sign motions that more faithfully convey the meaning of the input text.

\subsection{Ablation Studies}
\label{subsec:ablation_studies}

\paragraph{Text--sign retrieval analysis.}
We evaluate whether SeRV learns a more text-aligned token space via bidirectional text--sign retrieval, representing text with frozen T5 embeddings and motion with pooled latents from the frozen tokenizer encoder.
As shown in Table~\ref{tab:semantic_alignment_retrieval}, SeRV improves all recall metrics in both directions on How2Sign and OpenASL, indicating that semantic supervision improves cross-modal alignment.
\begin{table}[t]
\centering
\caption{Bidirectional text--sign retrieval with batch size 64 (higher is better).}
\label{tab:semantic_alignment_retrieval}
\scriptsize
\setlength{\tabcolsep}{3.2pt}
\renewcommand{\arraystretch}{0.92}
\begin{tabular}{llcccc}
\toprule
Dataset & Tokenizer & Direction & R@1$\uparrow$ & R@5$\uparrow$ & R@10$\uparrow$ \\
\midrule
\multirow{4}{*}{How2Sign}
& RVQ  & \multirow{2}{*}{T $\rightarrow$ S} & 11.61 & 37.28 & 59.04 \\
& \textbf{SeRV} &                                      & 14.52 & 46.97 & 73.21 \\
& RVQ  & \multirow{2}{*}{S $\rightarrow$ T} & 6.73  & 27.05 & 47.10 \\
& \textbf{SeRV} &                                      & 8.42  & 34.08 & 58.41 \\
\midrule
\multirow{4}{*}{OpenASL}
& RVQ  & \multirow{2}{*}{T $\rightarrow$ S} & 16.84 & 50.16 & 76.00 \\
& \textbf{SeRV} &                                      & 22.38 & 65.21 & 88.22 \\
& RVQ  & \multirow{2}{*}{S $\rightarrow$ T} & 9.83  & 36.39 & 62.01 \\
& \textbf{SeRV} &                                      & 12.98 & 47.30 & 79.18 \\
\bottomrule
\end{tabular}
\end{table}

\paragraph{Effect of \methodName Tokenization.}
We ablate the tokenizer objectives with a fixed generator (Table~\ref{tab:ablation_tokenizer}; Figure~\ref{fig:additional_serv} in Appendix~\ref{app:ablation}).
On both How2Sign and OpenASL, adding $\mathcal{L}_{\mathrm{global}}$ or $\mathcal{L}_{\mathrm{tok}}$ to the reconstruction-only RVQ consistently improves generation, especially for hands, and combining both performs best on all metrics, indicating that the two objectives are complementary.
The per-token alignment analysis in Appendix~\ref{sec:token-ablation} further suggests that $\mathcal{L}_{\mathrm{global}}$ aligns captions with motions at the sentence level, whereas $\mathcal{L}_{\mathrm{tok}}$ strengthens word-level matches.

\begin{table}[t]
\centering
\caption{Effect of SeRV tokenizer objectives on How2Sign and OpenASL. We report DTW-JPE (lower is better) and B-4 (higher is better). All variants use the same generator. RVQ uses only $\mathcal{L}_{\mathrm{RVQ}}$; SeRV adds both semantic objectives.}
\label{tab:ablation_tokenizer}
\setlength{\tabcolsep}{3pt}
\resizebox{\columnwidth}{!}{
\begin{tabular}{lcccccc}
\toprule
\multirow{2}{*}{Tokenizer Objective}
& \multicolumn{3}{c}{How2Sign}
& \multicolumn{3}{c}{OpenASL} \\
\cmidrule(lr){2-4}
\cmidrule(lr){5-7}
& Body$\downarrow$
& Hands$\downarrow$
& B-4$\uparrow$
& Body$\downarrow$
& Hands$\downarrow$
& B-4$\uparrow$ \\
\midrule

RVQ
& 6.82 & 9.63 & 2.89
& 9.50 & 11.95 & 3.90 \\

RVQ $+\mathcal{L}_{\mathrm{global}}$
& 6.67 & 9.50 & 3.15
& 9.26 & 11.66 & 4.02 \\

RVQ $+\mathcal{L}_{\mathrm{tok}}$
& 6.53 & 9.31 & 3.37
& 9.09 & 11.42 & 4.34 \\

\textbf{SeRV}
& \textbf{6.44} & \textbf{9.16} & \textbf{3.67}
& \textbf{8.81} & \textbf{11.10} & \textbf{4.82} \\

\bottomrule
\end{tabular}
}
\end{table}

\paragraph{Effect of Hierarchical Generation.}
Using the same SeRV tokenizer, we compare Hierarchical GPT with a flat autoregressive (Flat AR) baseline that predicts all residual tokens as one sequence and a two-stage baseline that predicts coarse and residual tokens in separate stages.
As shown in Table~\ref{tab:ablation_generator}, Hierarchical GPT outperforms both baselines on body and hand motion on both datasets, showing that preserving the residual token hierarchy is key to accurate ASL generation.

\begin{table}[t]
\centering
\caption{
Ablation on generator architectures on How2Sign and OpenASL.
All variants use the same SeRV tokenizer.
Lower is better for DTW-JPE. }
\label{tab:ablation_generator}
\resizebox{\columnwidth}{!}{
\begin{tabular}{lcccc}
\toprule
\multirow{2}{*}{Generator}
& \multicolumn{2}{c}{How2Sign}
& \multicolumn{2}{c}{OpenASL} \\
\cmidrule(lr){2-3}
\cmidrule(lr){4-5}
& Body$\downarrow$
& Hands$\downarrow$
& Body$\downarrow$
& Hands$\downarrow$ \\
\midrule

Flat AR
& 7.73 & 10.41
& 13.43 & 15.33 \\

Two-stage
& 7.21 & 9.86
& 12.24 & 13.80 \\

\textbf{Hierarchical GPT}
& \textbf{6.44} & \textbf{9.16}
& \textbf{8.81} & \textbf{11.10} \\

\bottomrule
\end{tabular}
}
\end{table}

\paragraph{Effect of Cumulative-Latent Supervision.}
We further examine where to apply $\mathcal{L}_{\mathrm{tok}}$: the level-0 latent, the cumulative latent of levels 0--1, or the full cumulative latent.
As shown in Table~\ref{tab:cumulative_latent_supervision}, supervising more complete cumulative latents consistently improves both generation and text-to-sign retrieval, which aligns with Hierarchical GPT conditioning each refinement level on lower-level tokens (see Appendix~\ref{app:cumulative_latent_supervision}).

\begin{table}[t]
\centering
\caption{
Effect of the $\mathcal{L}_{\mathrm{tok}}$ supervision target on How2Sign. Each row trains a separate tokenizer; supervising more complete cumulative latents improves generation and retrieval.
}
\label{tab:cumulative_latent_supervision}
\small
\setlength{\tabcolsep}{6pt}
\renewcommand{\arraystretch}{1.05}
\begin{tabular}{lcc}
\toprule
Supervision target & DTW-JPE B/H$\downarrow$ & R@1$\uparrow$ \\
\midrule
Level 0 ($\mathbf{q}^{(0)}$) & 6.62 / 9.45 & 12.65 \\
Levels 0--1 ($\mathbf{q}^{(0)}+\mathbf{q}^{(1)}$) & 6.56 / 9.32 & 13.80 \\
Full latent ($\hat{\mathbf{z}}$) & \textbf{6.44 / 9.16} & \textbf{14.52} \\
\bottomrule
\end{tabular}
\end{table}

\paragraph{Cumulative Contribution of RVQ Levels.}
We also reconstruct motion and perform retrieval using the cumulative latent up to each RVQ level (Table~\ref{tab:rvq_level_contribution}).
level 0 alone captures most of the cross-modal retrieval signal, while later levels mainly improve geometric reconstruction with smaller retrieval gains, indicating a coarse-to-fine division of labor across the hierarchy (see Appendix~\ref{app:rvq_level_contribution}).

\begin{table}[t]
\centering
\caption{
Cumulative contribution of RVQ levels on How2Sign, using the same trained SeRV tokenizer. level 0 captures most of the retrieval signal, while later levels mainly improve reconstruction. R@1: text-to-sign / sign-to-text.
}
\label{tab:rvq_level_contribution}
\small
\setlength{\tabcolsep}{3.5pt}
\renewcommand{\arraystretch}{1.05}
\begin{tabular}{lcc}
\toprule
Representation & PA-MPJPE B/H$\downarrow$ & R@1$\uparrow$ \\
\midrule
Level 0 ($\mathbf{q}^{(0)}$) & 20.84 / 8.06 & 13.61 / 7.79 \\
Levels 0--1 ($\mathbf{q}^{(0)}+\mathbf{q}^{(1)}$) & 17.42 / 6.71 & 14.18 / 8.16 \\
Full latent ($\hat{\mathbf{z}}$) & \textbf{16.23 / 6.13} & \textbf{14.52 / 8.42} \\
\bottomrule
\end{tabular}
\end{table}

\section{Conclusion}
We present \methodName, a semantic-aligned RVQ tokenizer for text-to-3D ASL generation. Unlike reconstruction-only tokenizers, \methodName uses global-level and token-level motion--text alignment to learn residual tokens that are both precise and text-predictable. Building on \methodName, Hierarchical GPT predicts residual tokens over time and quantization depth while preserving the coarse-to-fine hierarchy. To support scalable training and evaluation, we construct a large-scale reconstructed 3D ASL motion--text benchmark from OpenASL. Experiments on How2Sign and OpenASL show state-of-the-art performance, with consistent gains in body and hand motion accuracy; ablations confirm the contributions of both semantic alignment and hierarchy-aligned generation.

\section{Limitations}
Our framework has several limitations.
First, autoregressive decoding generates tokens sequentially, limiting inference parallelism for long sequences; masked, semi-autoregressive, or diffusion-based decoding could improve speed while preserving temporal coherence.
Second, although semantic alignment improves hand motion, our objective does not explicitly supervise handshape, palm orientation, or sign-level articulatory categories; hand-specific objectives or finer hand tokenization could address this.
Third, our representation does not model facial expression and emotion, which are essential non-manual components of sign language; facial parameters and emotion-aware supervision could yield more expressive and natural avatars.
Fourth, our reconstructed OpenASL motion is not noise-free; fine-grained finger articulation may still be inaccurate under occlusion or fast motion, and hands that leave the camera frame cannot be reliably recovered.
Finally, our framework conditions only on text; multi-modal inputs such as speech, glosses, reference videos, or style prompts could enable more controllable and expressive generation.

\section{Ethical Considerations}
We plan to publicly release the reconstructed OpenASL benchmark, subject to applicable permissions and licensing requirements. Any release would include only SMPL-X parameters indexed by the original OpenASL clip IDs, without redistributing the source videos (Appendix~\ref{app:data_release}).
Both human studies in this paper were approved by our institution's IRB, and all participants provided informed consent (Appendices~\ref{app:reconstruction_audit} and~\ref{app:user_study_details}).

\bibliography{custom}

@inproceedings{yu2024signavatars,
  title={Signavatars: A large-scale 3d sign language holistic motion dataset and benchmark},
  author={Yu, Zhengdi and Huang, Shaoli and Cheng, Yongkang and Birdal, Tolga},
  booktitle={European Conference on Computer Vision},
  pages={1--19},
  year={2024},
  organization={Springer}
}

@inproceedings{duarte2021how2sign,
  title={How2sign: a large-scale multimodal dataset for continuous american sign language},
  author={Duarte, Amanda and Palaskar, Shruti and Ventura, Lucas and Ghadiyaram, Deepti and DeHaan, Kenneth and Metze, Florian and Torres, Jordi and Giro-i-Nieto, Xavier},
  booktitle={Proceedings of the IEEE/CVF conference on computer vision and pattern recognition},
  pages={2735--2744},
  year={2021}
}

@inproceedings{shi2022open,
  title={Open-Domain Sign Language Translation Learned from Online Video},
  author={Shi, Bowen and Brentari, Diane and Shakhnarovich, Greg and Livescu, Karen},
  booktitle={Proceedings of the 2022 Conference on Empirical Methods in Natural Language Processing},
  pages={6365--6379},
  year={2022}
}

@article{saunders2021continuous,
  title={Continuous 3d multi-channel sign language production via progressive transformers and mixture density networks},
  author={Saunders, Ben and Camgoz, Necati Cihan and Bowden, Richard},
  journal={International journal of computer vision},
  volume={129},
  number={7},
  pages={2113--2135},
  year={2021},
  publisher={Springer}
}

@inproceedings{saunders2020progressive,
  title={Progressive transformers for end-to-end sign language production},
  author={Saunders, Ben and Camgoz, Necati Cihan and Bowden, Richard},
  booktitle={European Conference on Computer Vision},
  pages={687--705},
  year={2020},
  organization={Springer}
}

@inproceedings{jang2025lost,
  title={Lost in Translation, Found in Context: Sign Language Translation with Contextual Cues},
  author={Jang, Youngjoon and Raajesh, Haran and Momeni, Liliane and Varol, G{\"u}l and Zisserman, Andrew},
  booktitle={Proceedings of the IEEE/CVF Conference on Computer Vision and Pattern Recognition},
  pages={8742--8752},
  year={2025}
}

@inproceedings{wong2024sign2gpt,
  title={Sign2GPT: Leveraging Large Language Models for Gloss-Free Sign Language Translation},
  author={Wong, Ryan and Camgoz, Necati Cihan and Bowden, Richard},
  booktitle={International Conference on Learning Representations},
  year={2024}
}

@inproceedings{baltatzis2024neural,
  title={Neural Sign Actors: A Diffusion Model for 3D Sign Language Production from Text},
  author={Baltatzis, Vasileios and Potamias, Rolandos Alexandros and Ververas, Evangelos and Sun, Guanxiong and Deng, Jiankang and Zafeiriou, Stefanos},
  booktitle={Proceedings of the IEEE/CVF Conference on Computer Vision and Pattern Recognition},
  pages={1985--1995},
  year={2024}
}

@inproceedings{fang2025signdiff,
  title={Signdiff: Diffusion model for american sign language production},
  author={Fang, Sen and Sui, Chunyu and Zhou, Yanghao and Zhang, Xuedong and Zhong, Hongbin and Tian, Yapeng and Chen, Chen},
  booktitle={2025 IEEE 19th International Conference on Automatic Face and Gesture Recognition (FG)},
  pages={1--11},
  year={2025},
  organization={IEEE}
}

@inproceedings{qi2024signgen,
  title={Signgen: End-to-end sign language video generation with latent diffusion},
  author={Qi, Fan and Duan, Yu and Zhang, Huaiwen and Xu, Changsheng},
  booktitle={European Conference on Computer Vision},
  pages={252--270},
  year={2024},
  organization={Springer}
}

@inproceedings{yin2024t2s,
  title={T2S-GPT: Dynamic vector quantization for autoregressive sign language production from text},
  author={Yin, Aoxiong and Li, Haoyuan and Shen, Kai and Tang, Siliang and Zhuang, Yueting},
  booktitle={Proceedings of the 62nd Annual Meeting of the Association for Computational Linguistics (Volume 1: Long Papers)},
  pages={3345--3356},
  year={2024}
}

@inproceedings{zuo2025signs,
  title={Signs as tokens: A retrieval-enhanced multilingual sign language generator},
  author={Zuo, Ronglai and Potamias, Rolandos Alexandros and Ververas, Evangelos and Deng, Jiankang and Zafeiriou, Stefanos},
  booktitle={Proceedings of the IEEE/CVF International Conference on Computer Vision},
  pages={23806--23816},
  year={2025}
}

@inproceedings{lee2022autoregressive,
  title={Autoregressive image generation using residual quantization},
  author={Lee, Doyup and Kim, Chiheon and Kim, Saehoon and Cho, Minsu and Han, Wook-Shin},
  booktitle={Proceedings of the IEEE/CVF conference on computer vision and pattern recognition},
  pages={11523--11532},
  year={2022}
}

@inproceedings{saunders2022signing,
  title={Signing at scale: Learning to co-articulate signs for large-scale photo-realistic sign language production},
  author={Saunders, Ben and Camgoz, Necati Cihan and Bowden, Richard},
  booktitle={Proceedings of the IEEE/CVF Conference on Computer Vision and Pattern Recognition},
  pages={5141--5151},
  year={2022}
}

@inproceedings{fang2025signllm,
  title={Signllm: Sign language production large language models},
  author={Fang, Sen and Chen, Chen and Wang, Lei and Zheng, Ce and Sui, Chunyu and Tian, Yapeng},
  booktitle={2025 IEEE/CVF International Conference on Computer Vision Workshops (ICCVW)},
  pages={6681--6693},
  year={2025},
  organization={IEEE}
}

@article{jiang2023motiongpt,
  title={Motiongpt: Human motion as a foreign language},
  author={Jiang, Biao and Chen, Xin and Liu, Wen and Yu, Jingyi and Yu, Gang and Chen, Tao},
  journal={Advances in Neural Information Processing Systems},
  volume={36},
  pages={20067--20079},
  year={2023}
}

@inproceedings{guo2024momask,
  title={Momask: Generative masked modeling of 3d human motions},
  author={Guo, Chuan and Mu, Yuxuan and Javed, Muhammad Gohar and Wang, Sen and Cheng, Li},
  booktitle={Proceedings of the IEEE/CVF Conference on Computer Vision and Pattern Recognition},
  pages={1900--1910},
  year={2024}
}

@article{li2025uni,
  title={Uni-sign: Toward unified sign language understanding at scale},
  author={Li, Zecheng and Zhou, Wengang and Zhao, Weichao and Wu, Kepeng and Hu, Hezhen and Li, Houqiang},
  journal={arXiv preprint arXiv:2501.15187},
  year={2025}
}

@article{zuo2026madis,
  title={MaDiS: Taming Masked Diffusion Language Models for Sign Language Generation},
  author={Zuo, Ronglai and Potamias, Rolandos Alexandros and Sun, Qi and Ververas, Evangelos and Deng, Jiankang and Zafeiriou, Stefanos},
  journal={arXiv preprint arXiv:2601.19577},
  year={2026}
}

@inproceedings{radford2021learning,
  title={Learning transferable visual models from natural language supervision},
  author={Radford, Alec and Kim, Jong Wook and Hallacy, Chris and Ramesh, Aditya and Goh, Gabriel and Agarwal, Sandhini and Sastry, Girish and Askell, Amanda and Mishkin, Pamela and Clark, Jack and Krueger, Gretchen and Sutskever, Ilya},
  booktitle={International conference on machine learning},
  pages={8748--8763},
  year={2021},
  organization={PmLR}
}

@article{guo2025snapmogen,
  title={Snapmogen: Human motion generation from expressive texts},
  author={Guo, Chuan and Hwang, Inwoo and Wang, Jian and Zhou, Bing},
  journal={Advances in Neural Information Processing Systems},
  volume={38},
  pages={99939--99955},
  year={2025}
}

@article{touvron2023llama,
  title={Llama: Open and efficient foundation language models},
  author={Touvron, Hugo and Lavril, Thibaut and Izacard, Gautier and Martinet, Xavier and Lachaux, Marie-Anne and Lacroix, Timoth{\'e}e and Rozi{\`e}re, Baptiste and Goyal, Naman and Hambro, Eric and Azhar, Faisal and Rodriguez, Aurelien and Joulin, Armand and Grave, Edouard 
  and Lample, Guillaume },
  journal={arXiv preprint arXiv:2302.13971},
  year={2023}
}

@inproceedings{xie2024g2p,
  title={G2p-ddm: Generating sign pose sequence from gloss sequence with discrete diffusion model},
  author={Xie, Pan and Zhang, Qipeng and Taiying, Peng and Tang, Hao and Du, Yao and Li, Zexian},
  booktitle={Proceedings of the AAAI Conference on Artificial Intelligence},
  volume={38},
  number={6},
  pages={6234--6242},
  year={2024}
}

@inproceedings{xie2024sign,
  title={Sign language production with latent motion transformer},
  author={Xie, Pan and Peng, Taiying and Du, Yao and Zhang, Qipeng},
  booktitle={Proceedings of the IEEE/CVF Winter Conference on Applications of Computer Vision},
  pages={3024--3034},
  year={2024}
}

@inproceedings{sun2024aios,
  title={Aios: All-in-one-stage expressive human pose and shape estimation},
  author={Sun, Qingping and Wang, Yanjun and Zeng, Ailing and Yin, Wanqi and Wei, Chen and Wang, Wenjia and Mei, Haiyi and Leung, Chi-Sing and Liu, Ziwei and Yang, Lei and Cai, Zhongang},
  booktitle={Proceedings of the IEEE/CVF conference on computer vision and pattern recognition},
  pages={1834--1843},
  year={2024}
}

@article{cai2019dtwnet,
  title={DTWNet: A dynamic time warping network},
  author={Cai, Xingyu and Xu, Tingyang and Yi, Jinfeng and Huang, Junzhou and Rajasekaran, Sanguthevar},
  journal={Advances in neural information processing systems},
  volume={32},
  year={2019}
}

@article{loshchilov2017decoupled,
  title={Decoupled weight decay regularization},
  author={Loshchilov, Ilya and Hutter, Frank},
  journal={arXiv preprint arXiv:1711.05101},
  year={2017}
}

@inproceedings{esser2021taming,
  title={Taming transformers for high-resolution image synthesis},
  author={Esser, Patrick and Rombach, Robin and Ommer, Bjorn},
  booktitle={Proceedings of the IEEE/CVF conference on computer vision and pattern recognition},
  pages={12873--12883},
  year={2021}
}

@inproceedings{zelinka2020neural,
  title={Neural sign language synthesis: Words are our glosses},
  author={Zelinka, Jan and Kanis, Jakub},
  booktitle={Proceedings of the IEEE/CVF winter conference on applications of computer vision},
  pages={3395--3403},
  year={2020}
}

@article{stoll2020text2sign,
  title={Text2Sign: towards sign language production using neural machine translation and generative adversarial networks},
  author={Stoll, Stephanie and Camgoz, Necati Cihan and Hadfield, Simon and Bowden, Richard},
  journal={International Journal of Computer Vision},
  volume={128},
  number={4},
  pages={891--908},
  year={2020},
  publisher={Springer}
}

@article{yao2021filip,
  title={Filip: Fine-grained interactive language-image pre-training},
  author={Yao, Lewei and Huang, Runhui and Hou, Lu and Lu, Guansong and Niu, Minzhe and Xu, Hang and Liang, Xiaodan and Li, Zhenguo and Jiang, Xin and Xu, Chunjing},
  journal={arXiv preprint arXiv:2111.07783},
  year={2021}
}

@inproceedings{potamias2025wilor,
  title={Wilor: End-to-end 3d hand localization and reconstruction in-the-wild},
  author={Potamias, Rolandos Alexandros and Zhang, Jinglei and Deng, Jiankang and Zafeiriou, Stefanos},
  booktitle={Proceedings of the Computer Vision and Pattern Recognition Conference},
  pages={12242--12254},
  year={2025}
}

@article{oord2018representation,
  title={Representation learning with contrastive predictive coding},
  author={Oord, Aaron van den and Li, Yazhe and Vinyals, Oriol},
  journal={arXiv preprint arXiv:1807.03748},
  year={2018}
}

@article{jiang2023rtmpose,
  title={Rtmpose: Real-time multi-person pose estimation based on mmpose},
  author={Jiang, Tao and Lu, Peng and Zhang, Li and Ma, Ningsheng and Han, Rui and Lyu, Chengqi and Li, Yining and Chen, Kai},
  journal={arXiv preprint arXiv:2303.07399},
  year={2023}
}

@article{chung2024scaling,
  title={Scaling instruction-finetuned language models},
  author={Chung, Hyung Won and Hou, Le and Longpre, Shayne and Zoph, Barret and Tay, Yi and Fedus, William and Li, Yunxuan and Wang, Xuezhi and Dehghani, Mostafa and Brahma, Siddhartha and others},
  journal={Journal of machine learning research},
  volume={25},
  number={70},
  pages={1--53},
  year={2024}
}

\appendix
\section{Appendix}

\subsection{Curating SMPL-X Poses}
\label{sec:pose_curation}

We curate SMPL-X annotations by fusing three complementary estimators: AiOS~\cite{sun2024aios} for whole-body pose, WiLoR~\cite{potamias2025wilor} for fine-grained hand articulation, and RTMPose~\cite{jiang2023rtmpose} for accurate 2D arm keypoints.
Given a signing video, we initialize SMPL-X parameters with AiOS, taking the highest-confidence person in each frame and stabilizing body shape and camera parameters across the sequence. We then replace the AiOS finger poses with WiLoR estimates, which better capture the handshapes that carry ASL meaning.

Since the independently estimated hands may be misaligned with the body, we refine the upper-body pose so that the arms connect consistently to the WiLoR hands while matching the 2D evidence. With the finger poses fixed, we optimize a per-frame correction $\delta_f$ to the initial upper-body pose by minimizing
\begin{equation}
\label{eq:curation_total}
\mathcal{L}_{\mathrm{fit}} = \mathcal{L}_{\mathrm{2D}} + \lambda_{\mathrm{ori}}\mathcal{L}_{\mathrm{ori}} + \lambda_{\mathrm{temp}}\mathcal{L}_{\mathrm{temp}} + \lambda_{\mathrm{reg}}\mathcal{L}_{\mathrm{reg}}.
\end{equation}
The reprojection term aligns the projected shoulder, elbow, and wrist joints with the RTMPose keypoints and WiLoR wrist positions:
\begin{equation}
\label{eq:curation_2d}
\mathcal{L}_{\mathrm{2D}} = \textstyle\sum_{f,j} w_{f,j}\, \rho\big(\Pi(\hat{J}_{f,j}) - J^{\mathrm{2D}}_{f,j}\big),
\end{equation}
where $\Pi$ is the perspective projection, $w_{f,j}$ is a per-joint confidence weight, and $\rho$ is a robust penalty. The orientation term $\mathcal{L}_{\mathrm{ori}}$ aligns the reconstructed wrist rotations with the WiLoR global hand orientations. The temporal term penalizes the velocity $\dot{\delta}_f = \delta_f - \delta_{f-1}$ and acceleration $\ddot{\delta}_f = \delta_f - 2\delta_{f-1} + \delta_{f-2}$ of the corrections:
\begin{equation}
\label{eq:curation_temp}
\mathcal{L}_{\mathrm{temp}} = \textstyle\sum_f \big(\|\dot{\delta}_f\|_2^2 + \lambda_{\mathrm{acc}}\|\ddot{\delta}_f\|_2^2\big),
\end{equation}
and the regularizer keeps the pose near its initialization and prevents arm--torso penetration:
\begin{equation}
\label{eq:curation_reg}
\mathcal{L}_{\mathrm{reg}} = \textstyle\sum_f \big(\|\delta_f\|_W^2 + \lambda_{\mathrm{col}}\mathcal{L}_{\mathrm{col},f}\big),
\end{equation}
where $\|\cdot\|_W$ is a per-joint weighted $\ell_2$ norm and $\mathcal{L}_{\mathrm{col},f}$ is the arm--torso collision penalty for frame $f$.
We optimize in two stages, first fitting the positional terms and then adding $\mathcal{L}_{\mathrm{ori}}$ to correct the wrist orientations.

\subsection{Additional Controlled Ablation Studies}
\label{app:additional_ablation}

We conduct additional controlled experiments to isolate the contributions of the semantic objectives and analyze the behavior of the residual representation hierarchy.
Unless otherwise specified, all variants share the same architecture, data split, training schedule, and evaluation protocol.

The analyses in Appendix~\ref{app:cumulative_latent_supervision} and Appendix~\ref{app:rvq_level_contribution} are summarized in Section~\ref{subsec:ablation_studies} of the main text, where the corresponding tables (Table~\ref{tab:cumulative_latent_supervision} and Table~\ref{tab:rvq_level_contribution}) are reported. The semantic-loss weight analysis in Appendix~\ref{app:loss_weight_sensitivity} is reported only in this appendix (Table~\ref{tab:loss_weight_sensitivity}). This appendix gives the full setup and extended discussion.

\begingroup
\subsubsection{Controlled Tokenizer-Objective Ablation}
\label{app:controlled_tokenizer_ablation}

Table~\ref{tab:controlled_tokenizer_ablation} consolidates the reconstruction and generation results of each tokenizer objective, extending Table~\ref{tab:ablation_tokenizer}.
The generation variants use the same generator and differ in whether the global-level alignment objective $\mathcal{L}_{\mathrm{global}}$ and the token-level text-conditioned objective $\mathcal{L}_{\mathrm{tok}}$ are enabled.
Adding either semantic objective to the reconstruction-only RVQ improves both generation and reconstruction on both datasets, and combining both performs best on all metrics, indicating that semantic supervision does not come at the cost of reconstruction fidelity.

\begin{table*}[t]
\centering
\caption{
Tokenizer-objective ablation on How2Sign and OpenASL.
Reconstruction is measured by PA-MPJPE (B/H, in mm) and generation by DTW-JPE (B/H; lower is better for both) and B-4 (higher is better).
All variants share the same RVQ architecture and generator. Each semantic objective improves both reconstruction and generation, and combining them performs best on all metrics.
}
\label{tab:controlled_tokenizer_ablation}
\small
\setlength{\tabcolsep}{4.5pt}
\renewcommand{\arraystretch}{1.05}
\resizebox{\textwidth}{!}{
\begin{tabular}{lcccc}
\toprule
Tokenizer
& H2S Recon. B/H$\downarrow$
& OpenASL Recon. B/H$\downarrow$
& H2S Gen. B/H$\downarrow$, B-4$\uparrow$
& OpenASL Gen. B/H$\downarrow$, B-4$\uparrow$ \\
\midrule
RVQ
& 17.65 / 6.85
& 18.92 / 6.41
& 6.82 / 9.63, 2.89
& 9.50 / 11.95, 3.90 \\
RVQ $+\mathcal{L}_{\mathrm{global}}$
& 17.05 / 6.55
& 18.31 / 6.12
& 6.67 / 9.50, 3.15
& 9.26 / 11.66, 4.02 \\
RVQ $+\mathcal{L}_{\mathrm{tok}}$
& 16.65 / 6.35
& 17.88 / 5.89
& 6.53 / 9.31, 3.37
& 9.09 / 11.42, 4.34 \\
\textbf{SeRV}
& \textbf{16.23 / 6.13}
& \textbf{17.53 / 5.65}
& \textbf{6.44 / 9.16, 3.67}
& \textbf{8.81 / 11.10, 4.82} \\
\bottomrule
\end{tabular}
}
\end{table*}

On both datasets, adding $\mathcal{L}_{\mathrm{global}}$ or $\mathcal{L}_{\mathrm{tok}}$ to the reconstruction-only RVQ lowers body/hand DTW-JPE and raises B-4, e.g., from 6.82/9.63 to 6.67/9.50 and 6.53/9.31 on How2Sign, and from 9.50/11.95 to 9.26/11.66 and 9.09/11.42 on OpenASL.
Combining both objectives gives the best generation results on both datasets (6.44/9.16 and 8.81/11.10).
Reconstruction error also decreases as the semantic objectives are added, from 17.65/6.85 to 16.23/6.13 on How2Sign and from 18.92/6.41 to 17.53/5.65 on OpenASL, indicating that semantic supervision does not compromise reconstruction fidelity.
\par
\endgroup

\subsubsection{Effect of Cumulative-Latent Supervision}
\label{app:cumulative_latent_supervision}

We further investigate which residual representation should receive token-level semantic supervision.
Let $\mathbf{q}^{(l)}$ denote the quantized contribution of RVQ level $l$, using the notation from Section~\ref{sec:preliminaries}.
The cumulative latent up to level $k\in\{0,\ldots,L-1\}$ is defined as
\begin{equation}
\hat{\mathbf{z}}^{(\leq k)}
=
\sum_{l=0}^{k}\mathbf{q}^{(l)}.
\end{equation}

We train otherwise identical tokenizers by applying $\mathcal{L}_{\mathrm{tok}}$ to the level-0 latent, the cumulative latent from levels 0--1, or the full cumulative latent from all RVQ levels.

As shown in Table~\ref{tab:cumulative_latent_supervision}, applying supervision to increasingly complete cumulative representations consistently improves semantic retrieval and generation quality.
Body/hand DTW-JPE decreases from 6.62/9.45 with level-0 supervision to 6.56/9.32 with levels 0--1 and 6.44/9.16 with the full latent.
R@1 similarly increases from 12.65 to 13.80 and 14.52.

This result is consistent with the design of the Hierarchical GPT.
Each refinement level is generated conditioned on the text and lower-level motion tokens.
Supervising the cumulative latent therefore encourages the representation available at each refinement stage to remain compatible with this conditional generation process.

This does not imply that later residual levels independently encode word-level semantics.

\subsubsection{Cumulative Contribution of RVQ Levels}
\label{app:rvq_level_contribution}

To characterize the role of each residual level, we evaluate progressively accumulated latent representations from the trained SeRV tokenizer.
For each setting, we reconstruct motion and evaluate text-to-sign and sign-to-text retrieval using the cumulative latent available up to the corresponding RVQ level.

Table~\ref{tab:rvq_level_contribution} shows that level 0 already captures most of the cross-modal retrieval signal, achieving T2S and S2T R@1 scores of 13.61 and 7.79, respectively.
Adding subsequent residual levels substantially improves reconstruction, reducing body and hand PA-MPJPE from $20.84/8.06$ to $16.23/6.13$.
Retrieval performance also improves gradually to $14.52/8.42$.

These results indicate a coarse-to-fine division of labor across the RVQ hierarchy.
The first level represents the dominant motion structure and most of the text-related information, whereas later levels mainly refine geometric details while slightly strengthening cross-modal alignment.

\subsubsection{Semantic-Loss Weight Sensitivity}
\label{app:loss_weight_sensitivity}

\begin{table}[t]
\centering
\caption{
Sensitivity to the semantic-loss weight on How2Sign (PA-MPJPE in mm). The reported weight of $0.005$ gives the best generation; a smaller weight slightly favors reconstruction, while a larger one degrades both.
}
\label{tab:loss_weight_sensitivity}
\small
\setlength{\tabcolsep}{3pt}
\renewcommand{\arraystretch}{1.05}
\begin{tabular}{ccc}
\toprule
$\lambda_{\mathrm{sem}}$ & Recon. PA-MPJPE B/H$\downarrow$ & Gen. DTW-JPE B/H$\downarrow$ \\
\midrule
0.05 & 18.71 / 7.02 & 6.73 / 9.58 \\
0.005 & 16.23 / 6.13 & \textbf{6.44 / 9.16} \\
0.0005 & \textbf{15.77 / 6.05} & 6.66 / 9.40 \\
\bottomrule
\end{tabular}
\end{table}

We analyze the sensitivity of SeRV by tying the two semantic-loss weights in the SeRV training objective: $\lambda_{\mathrm{global}}=\lambda_{\mathrm{tok}}=\lambda_{\mathrm{sem}}$.
All model components and other training settings are fixed while jointly varying these two weights.

As shown in Table~\ref{tab:loss_weight_sensitivity}, a relatively small weight of $0.0005$ slightly favors reconstruction, achieving body and hand PA-MPJPE of $15.77/6.05$.
However, it produces weaker downstream generation than the reported setting.
The weight $\lambda_{\mathrm{sem}}=0.005$ achieves the best generation performance, with body and hand DTW-JPE of $6.44/9.16$.
Increasing the weight to $0.05$ degrades both reconstruction and generation.

The results reveal a trade-off between geometric reconstruction and semantic predictability.
Among the weights tested in Table~\ref{tab:loss_weight_sensitivity}, $\lambda_{\mathrm{sem}}=0.005$ gives the lowest generation error, whereas $0.0005$ gives the lowest reconstruction error. The setting $\lambda_{\mathrm{global}}=\lambda_{\mathrm{tok}}=0.005$ matches the main experiments and yields the same body/hand DTW-JPE as Table~\ref{tab:main_results}.

\subsection{Per-Token Alignment Analysis}
\label{sec:token-ablation}

To examine how each semantic objective shapes the token space at the word level, we compare four tokenizers that differ only in their semantic losses: RVQ, RVQ $+\mathcal{L}_{\mathrm{global}}$, RVQ $+\mathcal{L}_{\mathrm{tok}}$, and SeRV.
Since tokenizers trained without $\mathcal{L}_{\mathrm{tok}}$ lack a word--motion projection, we freeze each tokenizer and train the same projection heads with the token-level objective, so that the differences reflect the quantized motion space.
For each test caption, we compute the cosine similarity between every word, averaged over its sub-words, and every quantized motion token of the paired signing.
Each word is summarized by its match margin, defined as its peak similarity on the paired motion minus its mean peak similarity over eight length-matched sequences from other videos.

\begin{figure}[t]
  \centering
  \includegraphics[width=\columnwidth]{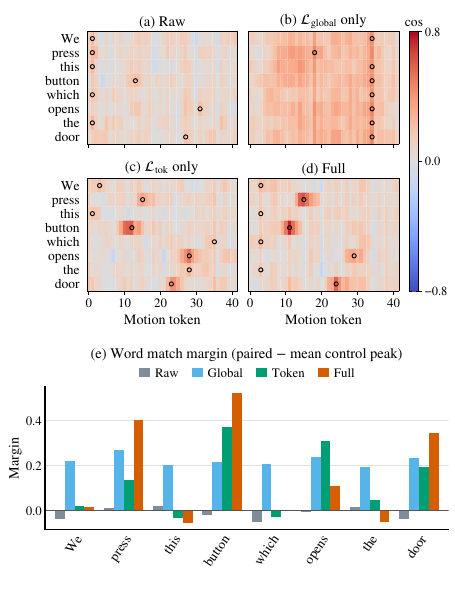}
  \caption{Per-token alignment of the test caption ``We press this button, which opens the door.'' under four training objectives. (a)--(d)~Word--token cosine similarity, where circles mark the best-matching token of each word. (e)~Word match margin.}
  \label{fig:abl-press}
\end{figure}

\begin{figure*}[t]
  \centering
  \begin{minipage}[t]{0.49\textwidth}
    \centering
    \includegraphics[width=\linewidth]{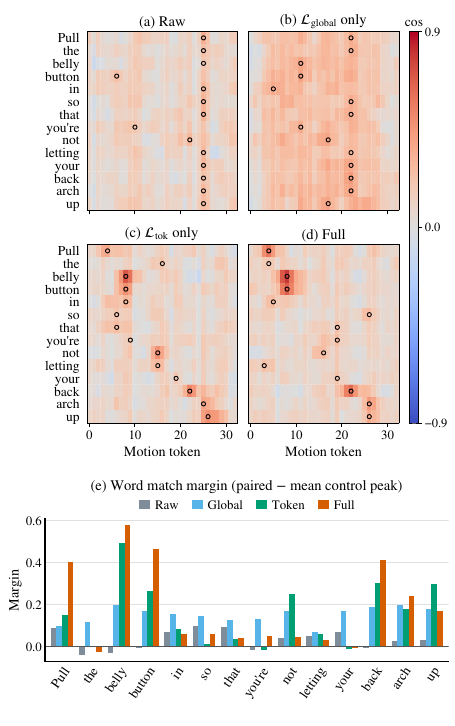}\\[2pt]
    {\small\textit{``Pull the belly button in so that you're not letting your back arch up.''}}
  \end{minipage}\hfill
  \begin{minipage}[t]{0.49\textwidth}
    \centering
    \includegraphics[width=\linewidth]{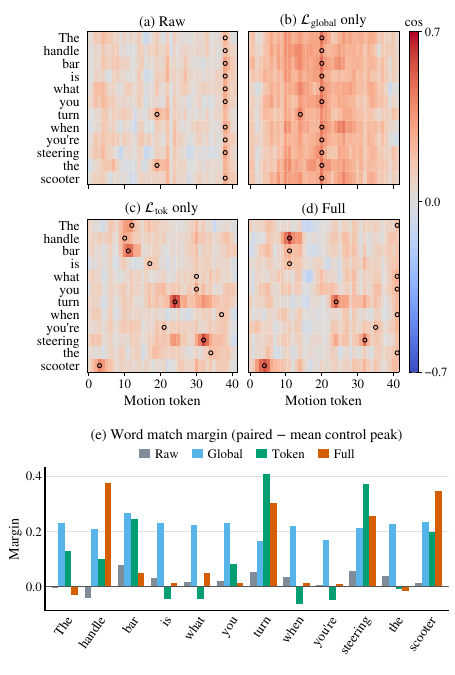}\\[2pt]
    {\small\textit{``The handle bar is what you turn when you're steering the scooter.''}}
  \end{minipage}
  \caption{Per-token alignment for two longer test captions, with panels as in Figure~\ref{fig:abl-press}.
  Under the full objective, \textit{belly button}, \textit{arch up}, and \textit{handle bar} each share a single segment, and \textit{scooter} is matched near the start of the signing although it ends the caption.}
  \label{fig:abl-long}
\end{figure*}

Figures~\ref{fig:abl-press} and~\ref{fig:abl-long} show three test captions.
With reconstruction-only latents, most words peak on the same generic motion token, and the margins remain close to zero.
$\mathcal{L}_{\mathrm{global}}$ alone raises the similarity of every word over a broad span of the signing but moves nearly all peaks to a single shared token, capturing that the caption and the motion match as a whole rather than which part of the motion each word corresponds to.
$\mathcal{L}_{\mathrm{tok}}$ alone produces sharper, word-specific peaks for several content words, such as \textit{button}, \textit{steering}, and \textit{belly}.
SeRV produces the sharpest peaks and the largest content-word margins on average, reaching \tbd{0.52} for \textit{button} and \tbd{0.40} for \textit{press}, while function words without a manual counterpart in ASL, such as \textit{this} and \textit{the}, stay near zero.

Qualitatively, the matches do not simply follow the English word order.
Multi-word expressions such as \textit{belly button}, \textit{arch up}, and \textit{handle bar} each peak on a single motion segment, and \textit{bar} receives only a small margin; \textit{scooter}, which ends its caption, peaks near the start of the signing, and \textit{button} is matched before \textit{press}.
Some content words, such as \textit{opens} and \textit{not}, remain weakly matched.

Across all \tbd{2,308} test captions, the mean content-word margin increases from \tbd{0.03} for the reconstruction-only RVQ to \tbd{0.11}, \tbd{0.33}, and \tbd{0.38} with $\mathcal{L}_{\mathrm{global}}$, $\mathcal{L}_{\mathrm{tok}}$, and SeRV, respectively.

The two objectives are therefore complementary: $\mathcal{L}_{\mathrm{global}}$ organizes the sentence-level geometry, whereas $\mathcal{L}_{\mathrm{tok}}$ strengthens word-level cross-modal association.

We note that this analysis measures word-specific cross-modal association rather than verified temporal alignment.
The late-interaction score is invariant to the order of motion tokens, the match margin reflects discrimination between paired and unpaired motion, and T5 token features are contextualized by the whole sentence, so a peak need not coincide with the sign of the corresponding word.
Verifying temporal correspondence would require signer-annotated sign boundaries, which we leave to future work.

\subsection{OpenASL Reconstruction Quality Audit}
\label{app:openasl_data_audit}

\subsubsection{Human Audit Protocol}
\label{app:reconstruction_audit}

To assess the quality of the reconstructed 3D motion, we randomly sample 50 clips from the OpenASL pairs and collect 300 independent ratings from human participants.
The participants are hearing volunteers recruited through a collaboration with our university. Participation was voluntary and unpaid; all participants provided informed consent, no personally identifying information was collected, and the audit was approved by our institution's Institutional Review Board (IRB).
Each reconstructed mesh sequence is shown side-by-side with its source video, and participants rate how well the reconstructed mesh reflects the signing in the source video on a five-point scale.
The five levels are defined as follows:
\begin{itemize}
\setlength{\itemsep}{0pt}
\item[1] \emph{Very poorly}: the reconstructed motion does not correspond to the signing in the video, and the hand and arm movements are largely incorrect.
\item[2] \emph{Poorly}: only parts of the signing are captured, with major errors in the hand or arm motion.
\item[3] \emph{Moderately}: the overall signing is recognizable, but noticeable errors remain in the handshapes, finger articulation, or movement trajectories.
\item[4] \emph{Well}: the reconstruction matches the signing with only minor deviations.
\item[5] \emph{Very well}: the reconstruction faithfully reproduces the signing, including fine-grained hand articulation.
\end{itemize}

\begin{table}[t]
\centering
\caption{
Human audit of reconstructed OpenASL motion.
Human participants rate 50 randomly sampled clips on a five-point scale, producing 300 ratings.
}
\label{tab:openasl_reconstruction_audit}
\small
\setlength{\tabcolsep}{3pt}
\renewcommand{\arraystretch}{1.05}
\begin{tabular}{llcc}
\toprule
Score & Rating & Number of ratings & Percentage \\
\midrule
5 & Very well & 180 & 60.0\% \\
4 & Well & 72 & 24.0\% \\
3 & Moderately & 30 & 10.0\% \\
2 & Poorly & 14 & 4.7\% \\
1 & Very poorly & 4 & 1.3\% \\
\midrule
\multicolumn{2}{l}{Total} & 300 & 100.0\% \\
\bottomrule
\end{tabular}
\end{table}

As shown in Table~\ref{tab:openasl_reconstruction_audit}, 84.0\% of the ratings judge the reconstruction to reflect the signing well or very well, 10.0\% moderately, and only 6.0\% poorly or very poorly, giving an average score of 4.37 out of 5.
These results indicate that most audited reconstructions faithfully capture the signing in their source videos.
Low scores mainly arise when the hands of the signer move out of the camera frame, where the reconstruction lacks visual evidence for the hands; these segments do not affect the meaning conveyed by the signing.
We keep all samples to preserve the official OpenASL split, and since all methods are trained and evaluated on the same reconstructions, this noise affects them equally.

\subsection{User Study Details}
\label{app:user_study_details}

This section provides details of the user study in Section~\ref{sec:user_study}.

\paragraph{Participants and recruitment.}
The study involved five professional ASL signers, recruited through a collaboration with our university.

\paragraph{Task and instructions.}
We randomly selected 20 sign sequences. For each sequence, participants were shown the input English text together with the avatars generated by \methodName, SOKE, and S-MotionGPT, and were asked to select the avatar that best conveyed the meaning of the text. Method names were hidden, and the display order of the methods was randomized.

\paragraph{Compensation.}
Participation was voluntary, and participants received no monetary compensation.

\paragraph{Consent and ethics.}
All participants provided informed consent and were told that their anonymized responses would be used for research publication. No personally identifying information was collected. The study was approved by our institution's Institutional Review Board (IRB).

\subsection{Comparison with Existing 3D Sign Resources}
\label{app:dataset_comparison}

Recent 3D sign language resources have made important progress toward avatar-based sign language understanding and generation. 
Early 3D ASL resources such as How2Sign provide aligned videos, depth, and English transcripts, but their scale and signer coverage remain limited for training high-capacity text-to-motion generators. 
More recent SLG systems further curate SMPL-X poses for existing continuous sign language benchmarks, including How2Sign, CSL-Daily, and Phoenix-2014T, enabling token-based 3D sign generation across multiple languages. 
In parallel, SignAvatars introduces a large-scale holistic 3D sign language dataset with SMPL-X, MANO, and 2D/3D keypoint annotations, covering both isolated and continuous signs with multiple prompt types. 
However, existing resources are either relatively limited in ASL scale, constructed from studio-recorded benchmarks, or designed for general multi-prompt/multilingual 3D sign production rather than large-scale paired text-to-3D ASL generation from in-the-wild videos.

Table~\ref{tab:dataset_comparison_3d} summarizes the comparison. 
We use the official OpenASL video--text pairs and data split without additional dataset filtering. OpenASL contains 98,417 translation pairs and 288 hours of ASL video from approximately 220 signers~\cite{shi2022open}, with 96,476 training, 966 validation, and 975 test pairs. We reconstruct SMPL-X motion for these paired clips. The duration denotes the official source-video corpus duration. 

\begin{table*}[t]
\centering
\caption{
Comparison with existing 3D sign language motion resources.
OpenASL pair counts and duration follow the official video--text corpus; duration denotes source videos rather than a separately measured post-reconstruction total.
``Text paired'' indicates whether spoken-language text is paired with the 3D motion.
}
\label{tab:dataset_comparison_3d}
\small
\setlength{\tabcolsep}{5.5pt}
\renewcommand{\arraystretch}{1.05}
\resizebox{\textwidth}{!}{
\begin{tabular}{lcccc}
\toprule
Dataset
& Lang.
& Clips / Pairs
& Duration
& Text Paired \\
\midrule
How2Sign-3D~\cite{duarte2021how2sign}
& ASL
& 35K
& 80+h
& \checkmark \\
SOKE Curated 3D Poses~\cite{zuo2025signs}
& ASL / CSL / DGS
& 35K / 20K / 8K
& --
& \checkmark \\
SignAvatars~\cite{yu2024signavatars}
& Multi-SL
& 70K
& 117h
& \checkmark \\
\midrule
Reconstructed OpenASL (Ours)
& ASL
& 98,417
& 288h
& \checkmark \\
\bottomrule
\end{tabular}
}
\end{table*}

\paragraph{Release and license.}
\label{app:data_release}
OpenASL is distributed under the CC BY-NC-ND 4.0 license~\cite{shi2022open}. We plan to publicly release the reconstructed OpenASL benchmark upon publication, subject to applicable permissions and licensing requirements. Any release would include only per-clip SMPL-X parameters indexed by the original OpenASL clip IDs. The source videos and English translations would not be redistributed and should be obtained from the original OpenASL release. The permissions and licensing terms applicable to distributing the reconstructed parameters remain to be confirmed.

\subsection{Evaluation Metric Definitions}
\label{subsec:metrics}

We provide formal definitions of the metrics used in Section~\ref{subsec:experimental_methodology}. PA-MPJPE is reported in millimeters (mm). DTW-JPE follows the evaluation protocol of SOKE~\cite{zuo2025signs}, so that our scores are directly comparable with previously reported How2Sign results.

\paragraph{Notation.}
Let $\hat{\mathbf{P}}=\{\hat{\mathbf{p}}_i\}_{i=1}^{\hat{T}}$ denote the joint sequence extracted from a generated motion and $\mathbf{P}=\{\mathbf{p}_j\}_{j=1}^{T}$ the corresponding reference joint sequence, where each frame contains $J$ 3D joints and $\mathbf{p}_j^{(k)},\hat{\mathbf{p}}_i^{(k)}\in\mathbb{R}^{3}$ are the reference and generated positions of joint $k$, respectively. Each frame is a $J\times3$ joint-coordinate array.
The per-frame joint position error between generated frame $i$ and reference frame $j$ is computed after translating the generated frame so that its alignment joint $r$ coincides with that of the reference frame:
\begin{equation}
d(i,j)=\frac{1}{J}\sum_{k=1}^{J}\bigl\|(\hat{\mathbf{p}}_i^{(k)}-\hat{\mathbf{p}}_i^{(r)})-(\mathbf{p}_j^{(k)}-\mathbf{p}_j^{(r)})\bigr\|_{2},
\label{eq:framecost}
\end{equation}
where $r$ is the pelvis for the upper-body joints and the wrist for each hand.

\paragraph{DTW-JPE.}
Since the generated and reference sequences may differ in length, we align them with dynamic time warping.
Let $\mathcal{W}$ be the set of monotonic warping paths $\pi=\{(i,j)\}$ satisfying the standard boundary, monotonicity, and continuity constraints.
DTW selects the path with minimum cumulative cost,
\begin{equation}
\pi^{\star}=\arg\min_{\pi\in\mathcal{W}}\sum_{(i,j)\in\pi}d(i,j),
\end{equation}
and DTW-JPE is the accumulated cost along the optimal path:
\begin{equation}
\mathrm{DTW\text{-}JPE}=\sum_{(i,j)\in\pi^{\star}}d(i,j).
\end{equation}
Following SOKE, the accumulated cost is not normalized by the path length and is computed on SMPL-X joint coordinates in meters; the reported score is its average over all test sequences.
We compute Eq.~\eqref{eq:framecost} over the body, left-hand, and right-hand joint sets separately, and report Hands as the mean of the left- and right-hand errors.

\paragraph{PA-MPJPE.}
For tokenizer reconstruction the output is frame-aligned with the input ($\hat{T}=T$), so no temporal warping is required.
For each frame $t$, we apply a similarity Procrustes alignment that estimates a rotation $\mathbf{R}_t\in\mathrm{SO}(3)$, translation $\mathbf{a}_t\in\mathbb{R}^{3}$, and scale $s_t>0$. The optimal alignment minimizes the residual between the reconstructed and reference joints:
\begin{equation}
\begin{aligned}
(\mathbf{R}_t^{\star},\mathbf{a}_t^{\star},s_t^{\star})
&=\arg\min_{\mathbf{R}_t,\mathbf{a}_t,s_t}\\
&\quad\sum_{k=1}^{J}\bigl\|s_t\mathbf{R}_t\hat{\mathbf{p}}_t^{(k)}+\mathbf{a}_t-\mathbf{p}_t^{(k)}\bigr\|_{2}^{2}.
\end{aligned}
\end{equation}
PA-MPJPE is then the per-joint error averaged over all frames after alignment:
\begin{equation}
\begin{aligned}
\mathrm{PA\text{-}MPJPE}&=\frac{1}{TJ}\sum_{t=1}^{T}\sum_{k=1}^{J}\\
&\quad\bigl\|s_t^{\star}\mathbf{R}_t^{\star}\hat{\mathbf{p}}_t^{(k)}+\mathbf{a}_t^{\star}-\mathbf{p}_t^{(k)}\bigr\|_{2}.
\end{aligned}
\end{equation}

\paragraph{Back-translation BLEU-4.}
\label{app:back_translation_evaluator}
Our back-translation evaluator is based on Uni-Sign~\cite{li2025uni}, which takes 2D pose sequences as input. For each dataset, we reproject the ground-truth 3D motions of the training split into 2D poses and fine-tune the corresponding pretrained Uni-Sign model on these pose--text pairs. During evaluation, generated 3D motions are reprojected into 2D poses in the same way and translated by the frozen evaluator, and BLEU-4 is computed against the reference transcripts. All B-4 scores in Table~\ref{tab:main_results}, including baseline scores, are obtained through this re-evaluation.

For reference, Table~\ref{tab:ground_truth_back_translation} reports the back-translation BLEU-4 scores of ground-truth motions using the same dataset-specific evaluators and 3D-to-2D projection protocol.

\begin{table}[t]
\centering
\caption{Reference back-translation BLEU-4 scores for ground-truth motions on How2Sign and OpenASL.}
\label{tab:ground_truth_back_translation}
\small
\setlength{\tabcolsep}{6pt}
\begin{tabular}{lcc}
\toprule
Input & How2Sign & OpenASL \\
\midrule
Ground-truth motion & 8.2 & 12.4 \\
\bottomrule
\end{tabular}
\end{table}

\subsection{Implementation details.} 
\label{sec:implementaion_details}

We train the framework in two stages: SeRV tokenizer training followed by Hierarchical GPT training.
In the first stage, the SeRV tokenizer is trained on the training split with the RVQ reconstruction objective and two semantic alignment objectives: the global-level motion--text alignment loss and the token-level motion--text late-interaction loss.
Unless otherwise specified, we use $L=3$ residual quantization levels, a latent dimension of $C=512$, and a codebook size of $K=512$ at each level.
The motion encoder downsamples the input sequence by a factor of $4$, so each discrete token corresponds to four motion frames.

For semantic supervision, we use a frozen T5 encoder~\cite{chung2024scaling} to encode the paired transcript.
The mean-pooled T5 hidden states are used as sentence-level text embeddings for the global alignment loss, where they are aligned with the temporally pooled level-0 quantized motion representation.
The token-level T5 hidden states are used in the late-interaction loss, where they are matched with the fully quantized motion latent sequence.
The SeRV training objective is
$\mathcal{L}_{\mathrm{SeRV}}=\mathcal{L}_{\mathrm{RVQ}}+\lambda_{\mathrm{global}}\mathcal{L}_{\mathrm{global}}+\lambda_{\mathrm{tok}}\mathcal{L}_{\mathrm{tok}}$.
Unless otherwise specified, we set $\lambda_{\mathrm{global}}=\lambda_{\mathrm{tok}}=0.005$, $\beta=0.25$, and $\tau=\tau^{\prime}=0.07$. The sensitivity analysis in Appendix~\ref{app:loss_weight_sensitivity} ties these two weights and varies their common value $\lambda_{\mathrm{sem}}$.
We optimize the tokenizer with AdamW~\cite{loshchilov2017decoupled} for $500$ epochs using a batch size of $128$, an initial learning rate of $4\times10^{-4}$, a weight decay of $0.1$, and cosine learning-rate decay.

In the second stage, Hierarchical GPT is trained to autoregressively predict the residual token hierarchy produced by the frozen SeRV tokenizer.
Given frozen T5 text features, the generator predicts tokens over time and RVQ depth in a coarse-to-fine order.
This preserves the residual hierarchy learned by SeRV instead of flattening all residual tokens into a single sequence.

\paragraph{Coarse Motion Decoder.}
The level-0 decoder predicts coarse motion tokens from the text condition and previous temporal history.
Since the level-0 RVQ code carries the dominant motion structure, this decoder is responsible for generating the coarse text-conditioned signing trajectory.
We implement it as a causal Transformer with LLaMA-style blocks, including RMSNorm, SwiGLU feed-forward layers, rotary positional embeddings, and prefix-based conditioning on the T5 text features.
The T5 token features are projected to the generator hidden dimension and prepended as conditioning tokens, allowing the decoder to autoregressively predict coarse motion tokens while attending to the transcript representation.

\paragraph{Residual Refinement Decoder.}
Higher-level tokens are predicted by residual refinement decoders.
Each residual decoder predicts refinement tokens conditioned on the text feature, previous temporal tokens, and the lower-level tokens already generated at the current time step.
Rather than generating motion from scratch, these decoders refine the representation produced by lower RVQ levels.
Concretely, the level-1 decoder is conditioned on the level-0 token embeddings, while the level-2 decoder is conditioned on the additive coarse context formed by level-0 and level-1 token embeddings.
This design follows the additive structure of RVQ, where later codebooks encode residual details over the coarse motion state.

The generator uses a hidden dimension of $1024$, $16$ attention heads, a dropout rate of $0.1$, and a maximum token length of $100$.
The coarse motion decoder contains $9$ Transformer layers, while the two residual refinement decoders contain $6$ and $3$ layers, respectively.
We train the generator for $150$ epochs with AdamW, using a batch size of $16$, a learning rate of $2\times10^{-4}$, a weight decay of $0.01$, and cosine learning-rate decay.
At inference time, tokens are generated in the same coarse-to-fine order until an EOS token is predicted or the maximum length is reached.
We use autoregressive sampling with a temperature of $1.0$.
The predicted multi-level token streams are decoded by the frozen SeRV decoder into continuous SMPL-X motion features, which are then converted into joints or vertices for evaluation.
All experiments are conducted on four NVIDIA L40S GPUs; training the full framework (SeRV tokenizer and Hierarchical GPT) takes about one day, i.e., roughly 96 GPU hours.

\subsection{Additional Qualitative Results on Ablation Study}
\label{app:ablation}

\begin{figure}[t]
    \centering
    \includegraphics[width=\linewidth]{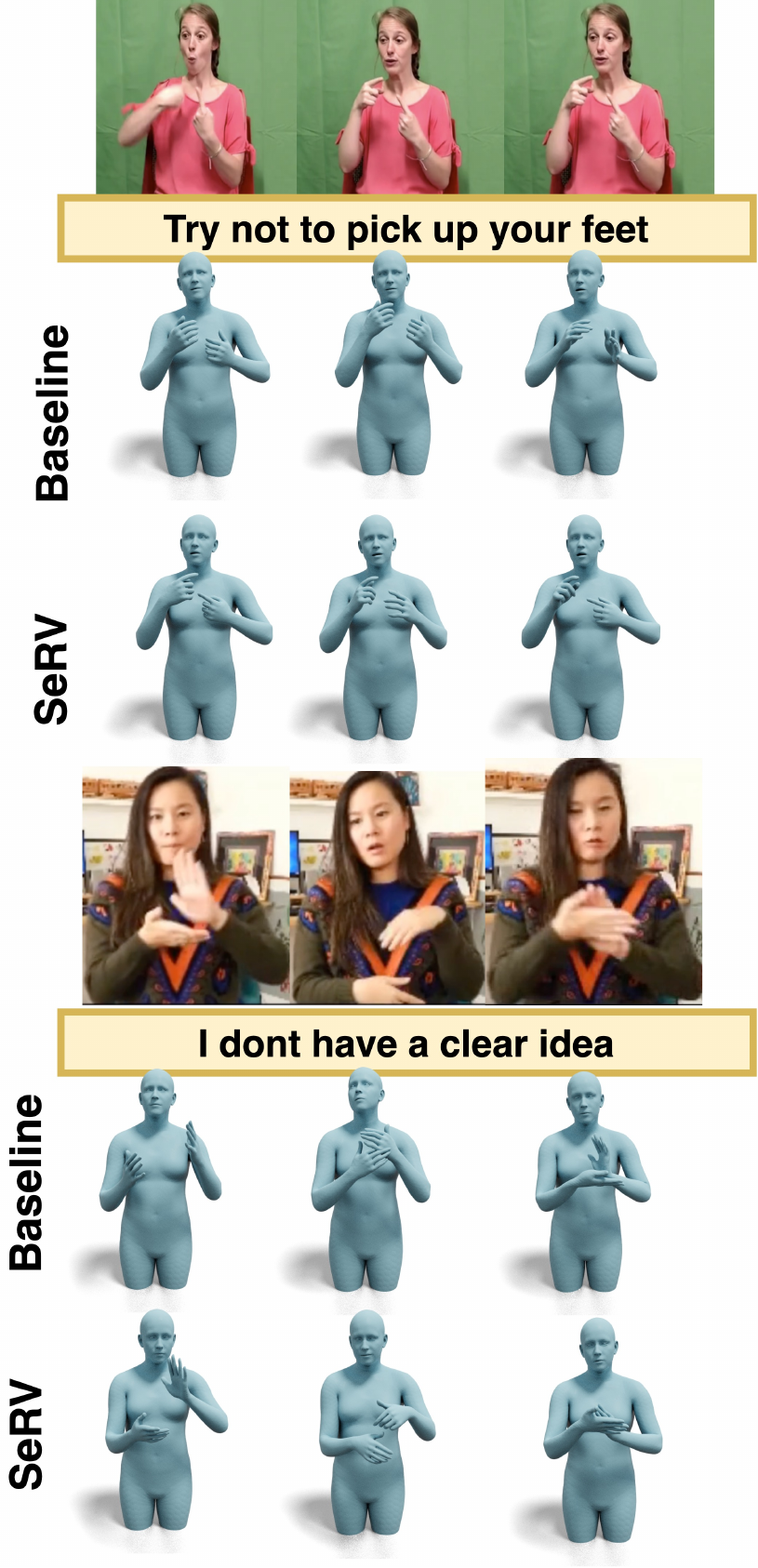}
    \caption{
Each example shows the reference (top), reconstruction-only baseline (middle, without $\mathcal{L}_{\mathrm{global}}$ and $\mathcal{L}_{\mathrm{tok}}$), and SeRV (bottom). Semantic-aligned RVQ enables SeRV to better preserve hand details and generate more faithful signing motions.
    }
    \label{fig:additional_serv}
\end{figure}

Figure~\ref{fig:additional_serv} provides additional qualitative ablation results, showing that semantic alignment improves hand articulation and the fidelity of generated 3D ASL motions.

\end{document}